\documentclass{article}

\PassOptionsToPackage{numbers,sort&compress}{natbib}

\usepackage[preprint]{neurips_2026}

\usepackage[utf8]{inputenc} 
\usepackage[T1]{fontenc}    
\usepackage{hyperref}       
\usepackage{url}            
\usepackage{booktabs}       
\usepackage{amsmath}        
\usepackage{amsfonts}       
\usepackage{nicefrac}       
\usepackage{microtype}      
\usepackage{xcolor}         
\usepackage{multirow}
\usepackage{subcaption}
\usepackage{graphicx}
\newcommand{\methodShort}{MISA}
\title{\methodShort{}: Mixture of Indexer Sparse Attention for Long-Context LLM Inference}

\author{Ruijie Zhou, Fanxu Meng\footnotemark[1], Yufei Xu, Tongxuan Liu, Guangming Lu, Muhan Zhang, Wenjie Pei\thanks{Corresponding author: \texttt{fxmeng@stu.pku.edu.cn}, \texttt{wenjiecoder@outlook.com}.}\\
  \centering\url{https://github.com/MuLabPKU/TransArch}}

\begin{document}

\maketitle

\begin{abstract}
DeepSeek Sparse Attention (DSA) sets the state of the art for fine-grained inference-time sparse attention by introducing a learned token-wise indexer that scores every prefix token and selects the top-$k$ for the main attention. To remain expressive, the indexer uses $H^I$ query heads (e.g.\ $64$ on DeepSeek-V3.2) that share the same selected token set; this multi-head design is precisely what makes the indexer the dominant cost on long contexts. We propose \textbf{\methodShort{}} (\textbf{M}ixture of \textbf{I}ndexer \textbf{S}parse \textbf{A}ttention), a drop-in replacement for the DSA indexer that treats its $H^I$ heads as a pool of mixture-of-experts: a lightweight router uses cheap block-level statistics to pick a query-dependent subset of $h \ll H^I$ active heads, and only those heads run the heavy token-level scoring. This preserves the diversity of the original indexer pool while reducing the per-query cost from $\mathcal{O}(H^I L)$ to $\mathcal{O}(h L + H^I M)$ with $M = \lceil L/B \rceil \ll L$ pooled keys. Following HISA, we further introduce a hierarchical variant, \methodShort$^\dagger$, that uses the MoE-routed pass to keep an enlarged candidate set and then re-ranks it with the original DSA indexer to recover the final top-$k$ almost exactly. With $h = 8$ active heads and \emph{no additional training}, \methodShort{} matches the dense DSA indexer on LongBench across DeepSeek-V3.2 and GLM-5 while running with $8\times$ and $4\times$ fewer indexer heads, respectively, and outperforms HISA on average; it preserves fully green Needle-in-a-Haystack heatmaps up to 128K context and recovers more than $92\%$ of the tokens selected by the DSA indexer per layer. Our TileLang kernel delivers roughly a $3.82\times$ speedup over DSA's original indexer kernel on a single NVIDIA H200 GPU. These results show that indexer-head-axis routing is a practical and complementary axis of efficiency for fine-grained sparse attention, on top of the existing token-axis hierarchies.
\end{abstract}

\section{Introduction}
\begin{figure}[t]
    \centering
    \begin{subfigure}[t]{0.475\textwidth}
        \centering
        \includegraphics[width=\linewidth]{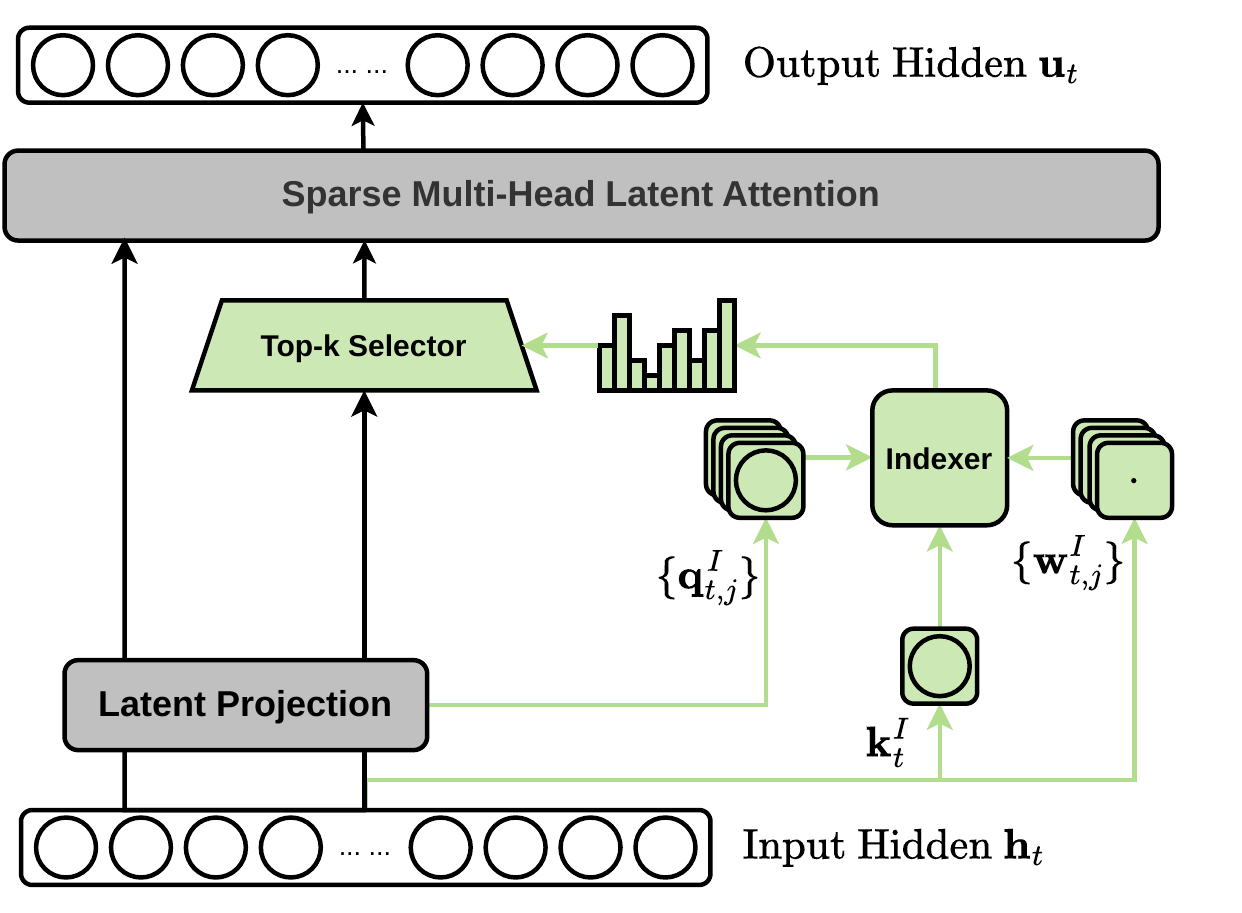}
        \caption{DSA}
        \label{fig:method_dsa}
    \end{subfigure}
    \hfill
    \begin{subfigure}[t]{0.475\textwidth}
        \centering
        \includegraphics[width=\linewidth]{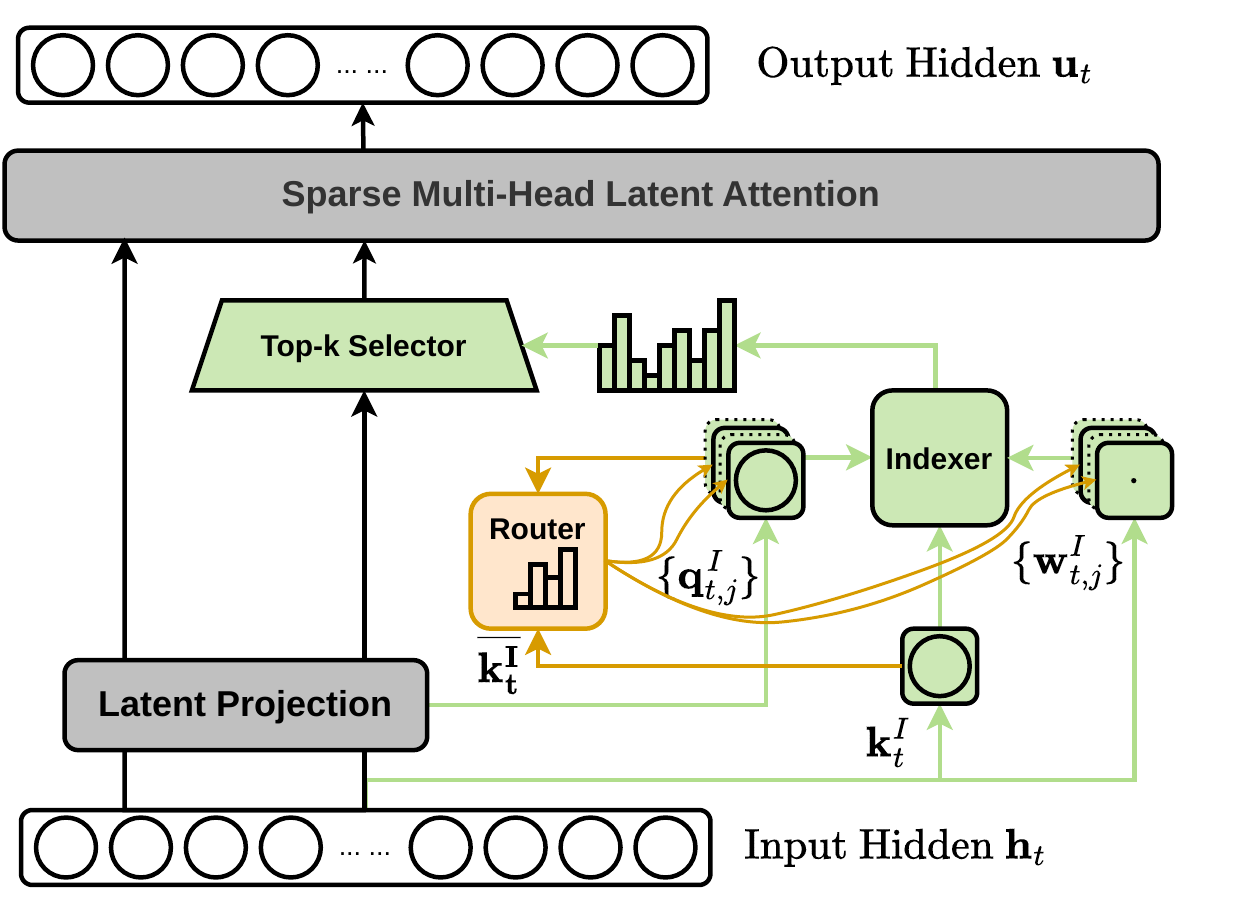}
        \caption{\methodShort}
        \label{fig:method_misa}
    \end{subfigure}
    \caption{Comparison of the DSA and \methodShort{} indexers. \textbf{(a) DSA} scores every prefix token with all $H^I$ indexer heads in parallel before the Top-$k$ Selector picks the final tokens. \textbf{(b) \methodShort{}} introduces a lightweight \emph{Router} that uses block-pooled indexing keys $\tilde{\mathbf{k}}_b^I$ (cf.\ Eq.~\ref{eq:pool}) to choose a query-dependent subset of $h \ll H^I$ active heads, and only those heads compute the per-token score. Both designs feed the same Sparse Multi-Head Latent Attention operator and produce a top-$k$ token set of identical size, so \methodShort{} is a drop-in replacement for the DSA indexer.}
    \label{fig:method}
\end{figure}

Frontier large language models such as Qwen3~\citep{yang2025qwen3}, Kimi~K2~\citep{moonshot2025kimik2}, GLM-5~\citep{zeng2026glm5}, and DeepSeek-V3.2~\citep{deepseekv32} now routinely process prefixes of hundreds of thousands of tokens in a single forward pass, and the latest releases---GPT-5.5~\citep{openai2026gpt55}, Claude~Opus~4.7~\citep{anthropic2026claudeopus47}, Gemini~3~\citep{google2026gemini3}, MiniMax-01~\citep{li2025minimax01}, and DeepSeek-V4~\citep{deepseekv4}---push this to the million-token regime. At these lengths, dense attention becomes the dominant cost of both prefill and decode, motivating a wave of \emph{sparse attention} techniques that select only a small subset $\mathcal{T}_t$ of past tokens for each query position~\citep{child2019sparse, beltagy2020longformer, zaheer2020bigbird, xiao2024streamingllm, zhang2023h2o, li2024snapkv, tang2024quest, xiao2024infllm, chen2024magicpig, gao2024seer, enzhe2025moba, yuan2025nsa, yufei2026hisa}. Among them, DeepSeek Sparse Attention (DSA)~\citep{deepseekv32} stands out as the best-performing fine-grained variant in production: rather than selecting whole blocks of tokens with handcrafted patterns, DSA introduces a lightweight learned \emph{indexer} that scores every prefix token and feeds the top-$k$ tokens into the main attention. Its dominance carries into the next generation: DeepSeek-V4's Compressed Sparse Attention (CSA)~\citep{deepseekv4} is, at its core, DSA applied on top of a $4\times$ compressed (block-level) KV stream, confirming that learned token-wise indexing remains the strongest building block even when the underlying KV is itself compressed. By contrast, block-level inference-time alternatives such as MoBA~\citep{enzhe2025moba} consistently lag behind DSA on retrieval-style benchmarks because their per-block scores cannot localise the relevant content within a block.

A central design choice of DSA is that the indexer itself is multi-head: although the main attention in DeepSeek-V3.2 has $H = 128$ query heads, all of them share the \emph{same} selected token set $\mathcal{T}_t$ (the sparse MLA operates in MQA mode with a single key/value entry per token), so a single indexer is sufficient---yet DSA still uses $H^I = 64$ indexer heads. The reason is expressiveness: each head specialises in a different relevance pattern (recency, syntactic role, lexical or semantic similarity, \ldots), and the aggregated score $I_{t,s} = \sum_{j} w_{t,j}^{I}\,\mathrm{ReLU}(\mathbf{q}_{t,j}^{I}\!\cdot\!\mathbf{k}_s^{I})$ benefits from this diversity, with measurable retrieval degradation as $H^I$ shrinks. The unfortunate consequence is that scoring each of the $L$ prefix tokens with all $H^I$ heads is precisely what makes the indexer the dominant cost on long contexts. Two recent improvements try to reduce this indexer cost without touching the head axis. IndexCache~\citep{bai2026indexcache} retains only a small set of layers as \emph{full} indexer layers and lets the remaining \emph{shared} layers reuse their top-$k$, amortising the cost \emph{across layers}; HISA~\citep{yufei2026hisa} attacks the cost from the \emph{token} axis with a hierarchical block-to-token search. Both still keep every one of the $H^I$ heads active inside the kernel, and both step away from DSA's strict per-token, per-layer scoring---tokens outside HISA's selected blocks and entire shared layers in IndexCache no longer receive a fresh fine-grained score, sacrificing part of the granularity that makes DSA strong.

This paper takes the orthogonal view that the bottleneck is the \emph{head} axis. Our key observation is that, while diversity across heads is essential when aggregated over a large pool, only a few heads are actually informative for any given query: the relevant set changes slowly along the prefix and can be identified from cheap block-level statistics. We turn this observation into \methodShort{} (\textbf{M}ixture of \textbf{I}ndexer \textbf{S}parse \textbf{A}ttention), an indexer that treats the $H^I$ heads as a pool of MoE experts, routes a query-dependent subset of $h \ll H^I$ active heads via a block-pooled scorer, and runs the heavy token-level scan with only those heads. The router itself operates on $M = \lceil L/B \rceil \ll L$ pooled keys, so its overhead is negligible, and the per-query indexer cost is reduced from $\mathcal{O}(H^I L)$ to $\mathcal{O}(h L + H^I M)$. \methodShort{} preserves the full diversity of the indexer pool because every head remains available; routing simply chooses \emph{which} ones to consult on each token. We further show that \methodShort{} can be plugged into a coarse-to-fine pipeline (\methodShort$^\dagger$) that uses an enlarged routed candidate set followed by a token-level DSA refinement, recovering the dense indexer's selections almost exactly.

\paragraph{Contributions.} (i) We identify the indexer's per-token head--token products as the dominant cost of DSA on long contexts and introduce head-axis routing as an axis of efficiency that is \emph{complementary} to the token-axis hierarchies explored by HISA~\citep{yufei2026hisa} and block-level methods~\citep{enzhe2025moba, tang2024quest, xiao2024infllm}. (ii) We propose \methodShort{}, a drop-in MoE-routed indexer that activates only $h \ll H^I$ heads per query through a lightweight block-pooled router, and a hierarchical extension \methodShort$^\dagger$ that re-ranks a routed candidate set with the original DSA indexer to recover the dense top-$k$ almost exactly. (iii) Without any additional training, \methodShort{} matches the dense DSA indexer on LongBench within $0.5$ average points across DeepSeek-V3.2 and GLM-5 while running with $h = 8$ active heads ($8\times$ and $4\times$ fewer indexer heads, respectively), preserves full Needle-in-a-Haystack accuracy up to 128K context, and recovers more than $92\%$ of the tokens selected by the DSA indexer per layer on \texttt{LSHT}. (iv) Our TileLang kernel implementation delivers roughly a $3.82\times$ speedup over DSA's original indexer kernel on a single NVIDIA H200 GPU. Together, these results show that head-level routing is a practical efficiency axis for fine-grained sparse attention, on top of any existing token-level scheme.

\section{Related work}
\paragraph{Sparse attention.} A long line of work attacks the quadratic cost of attention on long contexts by selecting a subset of past tokens for each query. \emph{Static-pattern} methods such as Sparse Transformer~\citep{child2019sparse}, Longformer~\citep{beltagy2020longformer}, and BigBird~\citep{zaheer2020bigbird} use predefined window, stride, and global tokens that are decoupled from the actual content. \emph{Cache-eviction} methods drop tokens at decode time using attention-statistics heuristics: StreamingLLM~\citep{xiao2024streamingllm} keeps a few attention sinks plus a recent window, H2O~\citep{zhang2023h2o} retains heavy hitters in past attention, and SnapKV~\citep{li2024snapkv} clusters and compresses the KV cache. These methods avoid retrieval entirely and can therefore lose information that becomes relevant later in the generation.

A more recent class of methods uses \emph{content-based dynamic retrieval} at the block level. Quest~\citep{tang2024quest} and InfLLM~\citep{xiao2024infllm} score blocks of tokens by their pooled key against the query and select the top-$m$ blocks at decode time. MoBA~\citep{enzhe2025moba} extends this idea to a trained inference-time selector with the same block-level granularity. MagicPIG~\citep{chen2024magicpig} replaces top-$k$ retrieval with LSH-based importance sampling. SeerAttention~\citep{gao2024seer} learns a sparse gate jointly with the model. Native Sparse Attention~\citep{yuan2025nsa} is the first work to train a transformer with a hardware-aligned sparsity pattern from scratch. Among inference-time methods, DeepSeek Sparse Attention~\citep{deepseekv32} is currently the strongest: a learned token-wise indexer scores every prefix token (rather than every block), and the resulting top-$k$ matches dense attention quality at production scale. HISA~\citep{yufei2026hisa} accelerates the DSA indexer with a coarse-to-fine block-to-token search, and IndexCache~\citep{bai2026indexcache} amortises the same indexer across consecutive layers by retaining only a few \emph{full} indexer layers and reusing their top-$k$ in the rest; both still keep every indexer head active inside the kernel and depart from per-token, per-layer scoring. \methodShort{} is complementary to all of these works: instead of acting along the token axis, we route along the head axis of the indexer itself, which can be combined orthogonally with any token-level scheme.

\paragraph{Mixture of experts in language models.} Conditional computation via mixture of experts (MoE) was introduced by~\citet{shazeer2017moe} and scaled up in GShard~\citep{lepikhin2020gshard} and Switch Transformer~\citep{fedus2022switch}, where a learned router activates a sparse subset of expert FFNs per token. Modern open-weight LLMs such as Mixtral~\citep{jiang2024mixtral} and DeepSeek-MoE~\citep{dai2024deepseekmoe} make their FFN layers MoE-based while leaving the attention layers dense. \emph{Attention-side} MoE has also been explored: Mixture-of-Attention-Heads~\citep{zhang2022moa} treats whole multi-head attention modules as experts and routes one module per token; MoH~\citep{jin2024moh} treats each individual attention head as an expert and selects a sparse subset of heads to compute the attention output. \methodShort{} adopts the same head-as-expert philosophy as MoH but applies it to a fundamentally different module. MoH and prior attention MoEs route the heads that produce the attention \emph{output}, so changing the routed subset directly changes the values written back into the residual stream. \methodShort{} instead routes the heads that produce the indexer \emph{score}: the routed subset only decides which similarity patterns are consulted when picking the top-$k$ tokens, while the downstream attention itself remains dense over the chosen set.This decoupling is what makes it possible to use a very small number of experts ($h \ll H^I$) without harming model quality.

\section{Preliminaries}
\label{sec:preliminaries}
We briefly review DeepSeek Sparse Attention (DSA) as used in DeepSeek-V3.2~\citep{deepseekv32} and a follow-up indexer-side improvement, HISA~\citep{yufei2026hisa}; both serve as baselines and as the starting points for the design of \methodShort{}. DSA consists of two components: a \textbf{token-wise indexer} that selects a small set of relevant tokens for each query, and \textbf{Sparse MLA} that performs attention only over the selected tokens. HISA introduces a \textbf{hierarchical indexing} mechanism that improves indexer efficiency while keeping Sparse MLA identical to DSA. The notation introduced below is used throughout the rest of the paper.

\paragraph{Indexer in DSA.}
Let $L$ denote the causal prefix length for a query position $t$. The indexer maintains lightweight indexing keys $\mathbf{k}_s^I$, indexing queries $\mathbf{q}_{t,j}^I$ for $H^I$ indexing heads, and per-head gating weights $w_{t,j}^I$. The relevance score between query $t$ and key $s$ is
\begin{equation}
I_{t,s} = \sum_{j=1}^{H^I} w_{t,j}^I \cdot \mathrm{ReLU}\!\left(\mathbf{q}_{t,j}^I \cdot \mathbf{k}_s^I\right).
\label{eq:dsa_score}
\end{equation}
The indexer then selects the top-$k$ token indices,
\begin{equation}
\mathcal{T}_t = \mathrm{TopK}(I_{t,:},\, k),
\label{eq:dsa_topk}
\end{equation}
which are passed to the downstream Sparse MLA operator. The per-query indexer cost is therefore $\mathcal{O}(H^I L)$, dominated by the $H^I$ head--token products in Eq.~\ref{eq:dsa_score}; aggregated over a full prefill it grows as $\mathcal{O}(H^I L^2)$ per layer.

\paragraph{Sparse MLA in DSA.}
Sparse MLA adopts the MQA mode of MLA, in which each token stores a single latent key--value entry $\mathbf{c}_s$ shared across all query heads. Given the selected token set $\mathcal{T}_t$, attention is computed only over the selected entries:
\begin{equation}
\mathbf{u}_t = \mathrm{Attn}\!\left(\mathbf{h}_t,\, \left\{\mathbf{c}_s \mid s \in \mathcal{T}_t\right\}\right).
\label{eq:sparse_mla}
\end{equation}
This reduces the dominant attention cost from $\mathcal{O}(L^2)$ to $\mathcal{O}(Lk)$. The interface between the two components is exactly the selected token set $\mathcal{T}_t$, which makes the indexer the natural target for further acceleration without touching Sparse MLA.

\paragraph{Indexer in HISA.}
HISA replaces the flat prefix scan with a two-stage coarse-to-fine search. The prefix is partitioned into $M = \lceil L / B \rceil$ contiguous blocks $\mathcal{B}_1, \ldots, \mathcal{B}_M$ of size $B$, and each block is summarized by a representative key obtained via mean pooling:
\begin{equation}
\tilde{\mathbf{k}}_b^I = \mathrm{Pool}\!\left(\left\{\mathbf{k}_s^I \mid s \in \mathcal{B}_b\right\}\right).
\label{eq:pool}
\end{equation}

In the first stage, the same indexing queries $\mathbf{q}_{t,j}^I$ and gating weights $w_{t,j}^I$ as in DSA are reused to score the pooled keys, and the top-$m$ blocks are selected:
\begin{equation}
J_{t,b} = \sum_{j=1}^{H^I} w_{t,j}^I \cdot \mathrm{ReLU}\!\left(\mathbf{q}_{t,j}^I \cdot \tilde{\mathbf{k}}_b^I\right),
\quad
\mathcal{C}_t = \mathrm{TopK}(J_{t,:},\, m).
\label{eq:block_score}
\end{equation}
Following~\citep{enzhe2025moba}, the first and last blocks are always included in $\mathcal{C}_t$ to retain the attention sink and local context. The candidate token set is then $\Omega_t = \bigcup_{b \in \mathcal{C}_t} \mathcal{B}_b$.

In the second stage, the original DSA scoring (Eq.~\ref{eq:dsa_score}) is applied within $\Omega_t$, and the final top-$k$ tokens
\begin{equation}
\mathcal{T}_t = \mathrm{TopK}\!\left(\left\{I_{t,s} \mid s \in \Omega_t\right\},\, k\right)
\label{eq:final_topk}
\end{equation}
are passed to the unmodified Sparse MLA operator.

\section{Method}
\label{method}

As established in Section~\ref{sec:preliminaries}, the DSA indexer scores every prefix token with all $H^I$ heads, giving a per-query cost of $\mathcal{O}(H^I L)$ that already dominates the indexer kernel even after FP8 quantisation, the ReLU non-linearity, and the small indexer dimension used by DeepSeek-V3.2~\citep{deepseekv32}. The $H^I$ heads cannot simply be collapsed into one: each specialises in a different relevance pattern, so reducing $H^I$ measurably degrades retrieval. Our starting point is the observation that this expressiveness is needed only \emph{in aggregate}---across all queries and across the prefix---whereas any single query is well served by a small, query-dependent subset of heads. The \methodShort{} method (Section~\ref{sec:method_misa}) exploits this by routing such a subset on cheap block-level statistics; a hierarchical extension \methodShort$^\dagger$ (Section~\ref{sec:method_hmisa}) re-introduces the full head pool only on a small candidate set, recovering DSA's exact selection.

\subsection{\methodShort{}: mixture of indexer experts}
\label{sec:method_misa}

We propose \methodShort{} (Mixture of Indexer Sparse Attention), shown alongside DSA in Figure~\ref{fig:method}. \methodShort{} treats the $H^I$ indexer heads as a pool of experts and uses a lightweight router to select a small subset of $h \ll H^I$ active heads per query, in the spirit of MoE routing. The selected experts then perform the token-level scoring, while the routing decision itself is computed on cheap block-level statistics.

\paragraph{Block-pooled router.}
Following Eq.~\ref{eq:pool}, the prefix is partitioned into $M = \lceil L/B \rceil$ blocks $\mathcal{B}_1, \ldots, \mathcal{B}_M$ and each block is summarized by a pooled indexing key $\tilde{\mathbf{k}}_b^I$. For query position $t$, the router computes per-head per-block affinities, weighted by the same gating coefficients $w_{t,j}^I$ used in the DSA score, and aggregates them across blocks to estimate the importance of each head:
\begin{equation}
A_{t,j,b} = w_{t,j}^I \cdot \mathrm{ReLU}\!\left(\mathbf{q}_{t,j}^I \cdot \tilde{\mathbf{k}}_b^I\right),
\qquad
E_{t,j} = \frac{1}{M}\sum_{b=1}^{M} \vert A_{t,j,b}\vert,
\label{eq:misa_route_score}
\end{equation}
and selects the top-$h$ heads as the active expert set:
\begin{equation}
\mathcal{H}_t = \mathrm{TopK}_j\!\left(E_{t,j},\, h\right).
\label{eq:misa_route}
\end{equation}
Including $w_{t,j}^I$ in $A_{t,j,b}$ makes $E_{t,j}$ a direct estimate of how much head $j$ would contribute to the final aggregated score $I_{t,s}$, rather than an unweighted similarity. The router operates on the $M \ll L$ pooled keys with all $H^I$ heads, so its cost is $\mathcal{O}(H^I M)$ per query. Crucially, since the router only needs to decide \emph{which heads are relevant} for the current query rather than which regions to keep, \methodShort{} can use a much coarser block partition than HISA: in our experiments, $B$ is set an order of magnitude larger than HISA's, which keeps $M$ small and makes the routing overhead negligible compared to the subsequent token-level scoring.

It is worth contrasting the role of block pooling here with that in HISA. Both methods compute the same per-head per-block affinities $A_{t,j,b}$ from pooled keys, but they reduce them along orthogonal axes: HISA aggregates \emph{across heads} to obtain a per-block score and selects the top-$m$ blocks, whereas \methodShort{} aggregates \emph{across blocks} to obtain a per-head importance and selects the top-$h$ heads. In other words, block pooling is used here purely as a cheap proxy that avoids materializing the full $\mathcal{O}(H^I L)$ tensor of head--token products when estimating which heads matter for the current query.

\paragraph{Sparse token scoring with active experts.}
Given $\mathcal{H}_t$, only the active heads compute the token-level score:
\begin{equation}
\hat{I}_{t,s} = \sum_{j \in \mathcal{H}_t} w_{t,j}^I \cdot \mathrm{ReLU}\!\left(\mathbf{q}_{t,j}^I \cdot \mathbf{k}_s^I\right),
\label{eq:misa_score}
\end{equation}
and the final token set
\begin{equation}
\mathcal{T}_t = \mathrm{TopK}_s\!\left(\hat{I}_{t,:},\, k\right)
\label{eq:misa_topk}
\end{equation}
is passed to the unmodified Sparse MLA operator. The per-query indexer cost is reduced from the DSA baseline $\mathcal{O}(H^I L)$ to $\mathcal{O}(H^I M + h L)$, where $M = \lceil L/B \rceil \ll L$ and $h \ll H^I$. In all of our experiments we use $h = 8$ on top of $H^I = 64$ (DeepSeek-V3.2) or $H^I = 32$ (GLM-5), so the dominant $h L$ term is reduced by $8\times$ or $4\times$ relative to DSA while the diversity of patterns available to the indexer is preserved---every head remains in the pool, and routing only chooses which ones to consult on each query.

\paragraph{Why MoE in the indexer.}
Standard MoE routes the FFN computation while keeping attention dense. \methodShort{} instead routes \emph{which similarity patterns are used to select tokens}; the downstream attention itself remains dense over the chosen $\mathcal{T}_t$. The key empirical observation that makes this practical is that the relevant indexer heads of a query change slowly across the prefix, so a coarse block-level estimate is sufficient to predict them, and the heavy token-level scan only needs to consult a small subset of heads.

\subsection{Hierarchical \methodShort{}: coarse-to-fine indexing}
\label{sec:method_hmisa}

The single-stage \methodShort{} above already cuts indexer cost substantially. We further extend it to a coarse-to-fine variant, denoted \methodShort$^\dagger$, in which a cheap MoE pass first prunes the prefix to a candidate set, and a second pass refines the candidates with the full DSA indexer. Crucially, the second stage of \methodShort$^\dagger$ uses the \emph{DSA indexer} (Eq.~\ref{eq:dsa_score}) on the candidate tokens; we do not adopt HISA's block-level top-$m$ filtering at any stage.

In the coarse stage, the MoE scoring of Eq.~\ref{eq:misa_score} is used to select an enlarged candidate set
\begin{equation}
\Omega_t = \mathrm{TopK}_s\!\left(\hat{I}_{t,:},\, k'\right), \quad k' > k.
\label{eq:misa_candidate}
\end{equation}
In the fine stage, the original DSA scoring is applied within $\Omega_t$ using all $H^I$ heads, and the final tokens
\begin{equation}
\mathcal{T}_t = \mathrm{TopK}_s\!\left(\left\{I_{t,s} \mid s \in \Omega_t\right\},\, k\right)
\label{eq:misa_final_topk}
\end{equation}
feed Sparse MLA.

\paragraph{Comparison with HISA.}
Both \methodShort$^\dagger$ and HISA have a two-stage coarse-to-fine structure, but the granularity of the coarse pass is fundamentally different. HISA's coarse pass operates at the block level: pooling discards intra-block variation and a block must be kept or discarded as a whole, so the candidate set $\Omega_t$ is always a union of full blocks. \methodShort$^\dagger$, by contrast, keeps the coarse pass at full token granularity (Eq.~\ref{eq:misa_candidate}) and reduces compute via head-level routing instead of block-level filtering, so tokens within the same block are still ranked individually. The fine pass is also stricter: it re-applies the original DSA scoring using all $H^I$ heads on $\Omega_t$, whereas HISA's fine pass operates on the union of selected blocks without revisiting the head set. Empirically, this combination yields a candidate set with consistently higher recall of DSA's top-$k$ at the same compute budget (Appendix~\ref{app:exp_iou}), which in turn translates into the quality gains reported in Sections~\ref{sec:exp_longbench}--\ref{sec:exp_niah}.

\section{Experimental results}
\label{exp}

We evaluate \methodShort{} as a drop-in replacement for the indexer in DeepSeek Sparse Attention (DSA), and verify that it preserves the retrieval and downstream quality of DSA / HISA at a fraction of the per-token compute. Unless stated otherwise, all sparse methods are applied at inference time without any additional training.

\paragraph{Models.} We use two open-weight long-context models that natively support DSA: \textbf{DeepSeek-V3.2}~\citep{deepseekv32} ($H^I = 64$ indexer heads, $H = 128$ main-attention heads, single shared KV head) and \textbf{GLM-5}~\citep{zeng2026glm5} ($H^I = 32$ indexer heads). All baselines (DSA, Block-Sparse, HISA) and all variants of \methodShort{} share the same Sparse Multi-Head Latent Attention operator and differ only in how the indexer produces the per-query token set $\mathcal{T}_t$.

\paragraph{Default \methodShort{} hyperparameters.} The router uses a block size of $B = 1024$, which is an order of magnitude larger than HISA's $B = 128$ and yields a small number of pooled keys $M = \lceil L/B \rceil$. The single-stage \methodShort{} routes $h$ active heads on the full prefix and selects $k = 2048$ tokens directly. The hierarchical \methodShort$^\dagger$ first uses the same MoE-routed scoring to retain $k' = 8192$ candidates, then re-scores them with the full DSA indexer (all $H^I$ heads) and keeps the top $k = 2048$. Unless otherwise specified, we use $h = 8$ active heads in both prefill and decode, i.e.\ a $8\times$ reduction over DSA on DeepSeek-V3.2 and a $4\times$ reduction on GLM-5.

\paragraph{Baselines.} \emph{DSA} is the original dense indexer of DeepSeek-V3.2 / GLM-5, scoring every prefix token with all $H^I$ indexer heads. \emph{Block-Sparse} is a MoBA-style~\citep{enzhe2025moba} inference-time selector: the prefix is partitioned into uniform blocks of size $B = 128$, each block is summarised by its mean indexing key, and the $m = k / B$ blocks with the highest query-block inner product are retained as the selected token set (same overall token budget $k$ as all other methods). \emph{HISA}~\citep{yufei2026hisa} is the hierarchical indexer with block size $B = 128$, top-$m$ block filtering, and per-token refinement. The DSA, Block-Sparse, and HISA scores in Table~\ref{tab:longbench} are taken from the HISA paper~\citep{yufei2026hisa}, while all other figures in this section are produced with our own implementation under matched hyperparameters.

\paragraph{Hardware and evaluation.} All experiments---LongBench evaluation, NIAH heatmaps, indexer-kernel benchmark, IoU computation, and the three ablation studies ---are run on 8$\times$ \textbf{NVIDIA H200} GPUs. We measure (i) downstream quality on \textbf{LongBench}, averaged across sub-tasks within each of six categories; (ii) \textbf{Needle-in-a-Haystack} retrieval accuracy at context lengths up to 128K, sweeping needle depth from $0\%$ to $100\%$; (iii)  1 and 2 stage \methodShort{}  \textbf{kernel latency} of the indexer; and (iv) per-layer \textbf{Intersection-over-Union (IoU)} between the token set selected by a given method and the top-$2048$ set selected by the full DSA indexer, which serves as the ground-truth reference for retrieval. Ablations on the router score $E_{t,j}$, the number of active heads $h$, and the block size $B$ are all carried out on DeepSeek-V3.2. The IoU experiments and the majority of ablation experiments are provided in the appendix.

\subsection{LongBench}
\label{sec:exp_longbench}
Table~\ref{tab:longbench} reports LongBench scores on DeepSeek-V3.2 and GLM-5. The DSA, Block-Sparse, and HISA rows are reproduced from the HISA paper~\citep{yufei2026hisa} and use the full $H^I$ indexer head pool ($H^I = 64$ on DeepSeek-V3.2 and $H^I = 32$ on GLM-5). \methodShort{} and \methodShort$^\dagger$ instead activate only $h = 8$ heads per query on \emph{both} models, which corresponds to using $1/8$ of the indexer heads on DeepSeek-V3.2 and $1/4$ on GLM-5; the final token budget is fixed at $k = 2048$ across all rows so that downstream attention sees the same workload.

Despite this much tighter head budget, \methodShort{} matches the dense DSA indexer on the average column on both models---within $0.20$ points on DeepSeek-V3.2 ($50.85$ vs.\ $51.05$) and surpassing it on GLM-5 ($46.43$ vs.\ $46.01$)---and outperforms Block-Sparse and HISA on average across both models. The hierarchical \methodShort$^\dagger$ further closes the residual gap on DeepSeek-V3.2 to $0.1$ average points ($50.95$ vs.\ $51.05$): every per-category score lands within $0.4$ points of DSA, and Single-Document QA actually improves marginally ($+0.02$). Effectively, \methodShort$^\dagger$ preserves the quality of DSA's full $H^I = 64$-head indexer using only $h = 8$ active heads. On GLM-5, where the head reduction over native DSA is $4\times$, both \methodShort{} variants beat DSA, Block-Sparse, and HISA on Multi-Doc QA, Summarisation, and Code, and \methodShort$^\dagger$ achieves the best overall average ($46.51$). The Block-Sparse baseline trails by $1.5$--$3.4$ average points on either model, confirming that block-uniform selection is too coarse for fine-grained retrieval-style tasks even when given the full head pool.

\begin{table}[htb]
\centering
\caption{LongBench results for DeepSeek-V3.2 and GLM-5 under different indexing strategies. All sparse methods are applied at inference time without additional training. The \textbf{Heads} column reports the number of active indexer heads used per query (out of $H^I = 64$ for DeepSeek-V3.2 and $32$ for GLM-5). Scores are averaged across sub-tasks within each category. Task abbreviations: \textbf{SQA}~=~Single-Document QA, \textbf{MQA}~=~Multi-Document QA, \textbf{Sum}~=~Summarization, \textbf{FS}~=~Few-shot Learning, \textbf{Syn}~=~Synthetic Retrieval, \textbf{Code}~=~Code Completion. Best per column in \textbf{bold}, second-best \underline{underlined}.}
\label{tab:longbench}
\vspace{0.5em}
\setlength{\tabcolsep}{5pt}
\begin{tabular}{ccc cccccc|c}
\toprule
\textbf{Model}  & \textbf{Indexer} & \textbf{Heads} & \textbf{SQA} & \textbf{MQA} & \textbf{Sum} & \textbf{FS} & \textbf{Syn}& \textbf{Code} & \textbf{Avg.} \\
\midrule
\multirow{5}{*}{DeepSeek-V3.2} 
& DSA & 64 & \underline{50.89} & \textbf{52.66} & 22.11 & \textbf{62.24} & 69.83 & 48.56 & \textbf{51.05}\\ 
& Block & 64 & 48.36 & 49.76 & 21.90 & 59.45 & 68.67 & \textbf{49.09} & 49.54 \\
& HISA & 64 & 49.17 & 51.96 & \underline{22.13} & 61.62 & \textbf{70.83} & \underline{48.99} & 50.78\\
& \methodShort & 8 & 50.83 & 51.62 & \textbf{22.31} & 61.77 & \underline{70.00} & 48.54 & 50.85 \\
& \methodShort$^\dagger$ & 8 & \textbf{50.91} & \underline{52.27} & 22.04 & \underline{62.10} & 69.83 & 48.45 & \underline{50.95} \\

\midrule
\multirow{5}{*}{GLM-5} 
& DSA & 32 & 41.23 & 27.89 & \underline{18.39} & 63.20 & 68.84 & 56.53 & 46.01 \\
& Block & 32 & 38.35 & 24.29 & 16.95 & 60.64 & 60.49 & 55.29 & 42.67 \\
& HISA & 32 & \textbf{42.45} & 27.62 & 17.90 & \textbf{63.78} & \textbf{69.35} & 56.79 & 46.32\\
& \methodShort & 8 & \underline{41.64} & \textbf{29.16} & 18.22 & 63.53 & 68.76 & \underline{57.24} & \underline{46.43} \\
& \methodShort$^\dagger$ & 8 & 41.16 & \underline{28.67} & \textbf{18.56} & \underline{63.57} & \underline{69.26} & \textbf{57.83} & \textbf{46.51} \\
\bottomrule
\end{tabular}
\end{table}

\subsection{Needle-in-a-Haystack retrieval accuracy}
\label{sec:exp_niah}
\begin{figure}[t]
    \centering
    \begin{subfigure}[t]{0.19\textwidth}
        \centering
        \includegraphics[width=\linewidth]{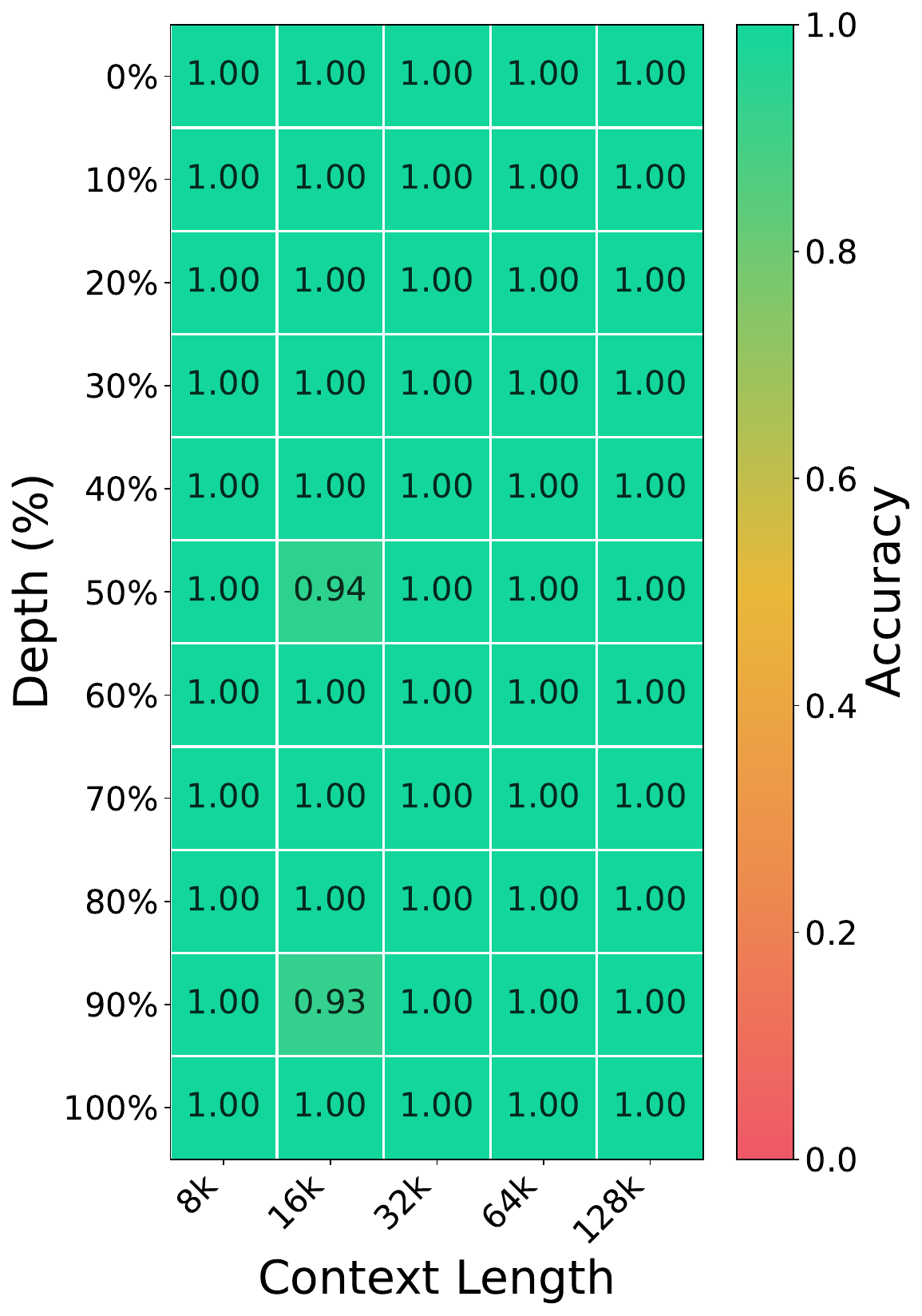}
        \caption{DSA (original)}
        \label{fig:niah_dsa}
    \end{subfigure}
    \hfill
    \begin{subfigure}[t]{0.19\textwidth}
        \centering
        \includegraphics[width=\linewidth]{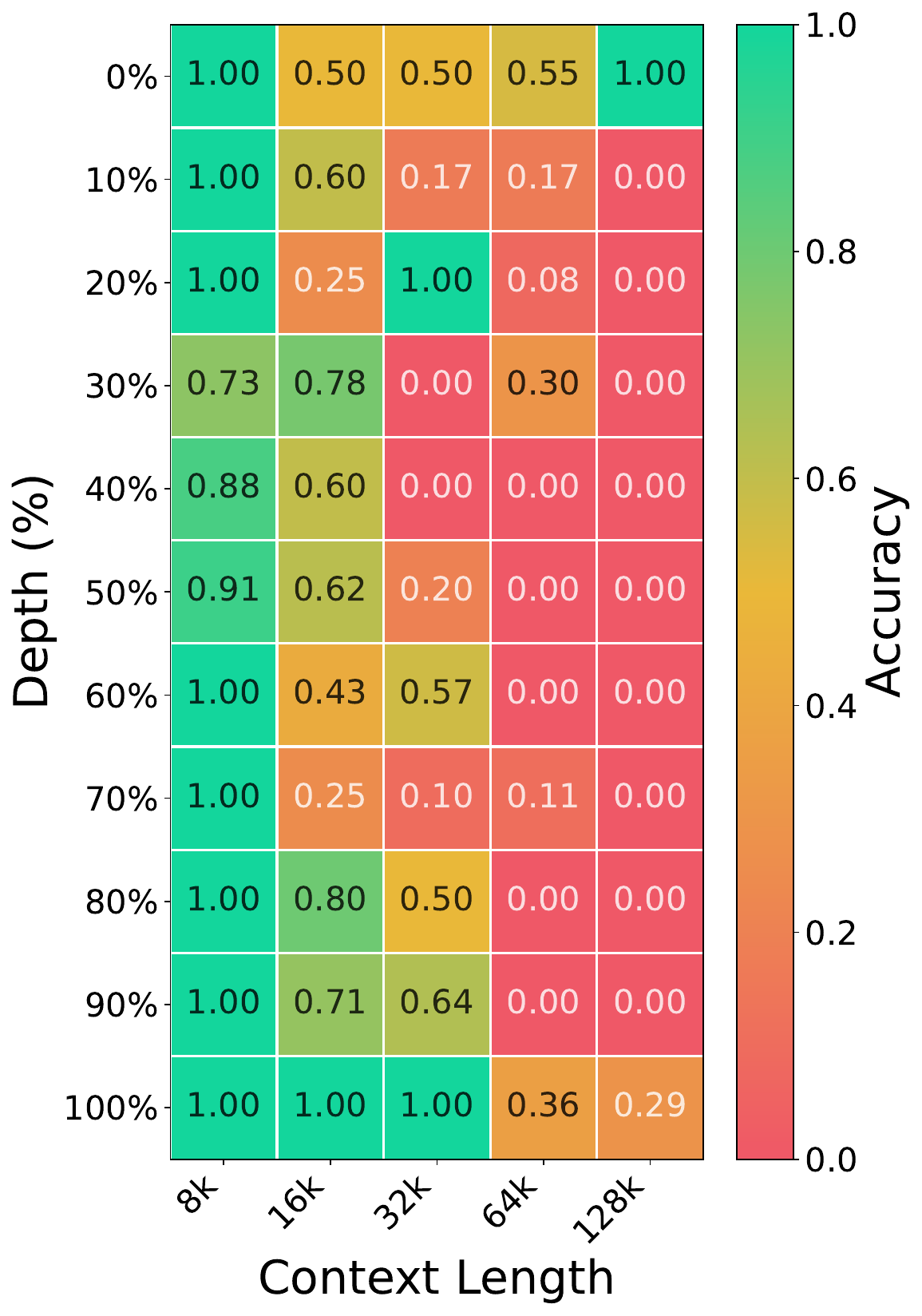}
        \caption{Block-Sparse}
        \label{fig:niah_block}
    \end{subfigure}
    \hfill
    \begin{subfigure}[t]{0.19\textwidth}
        \centering
        \includegraphics[width=\linewidth]{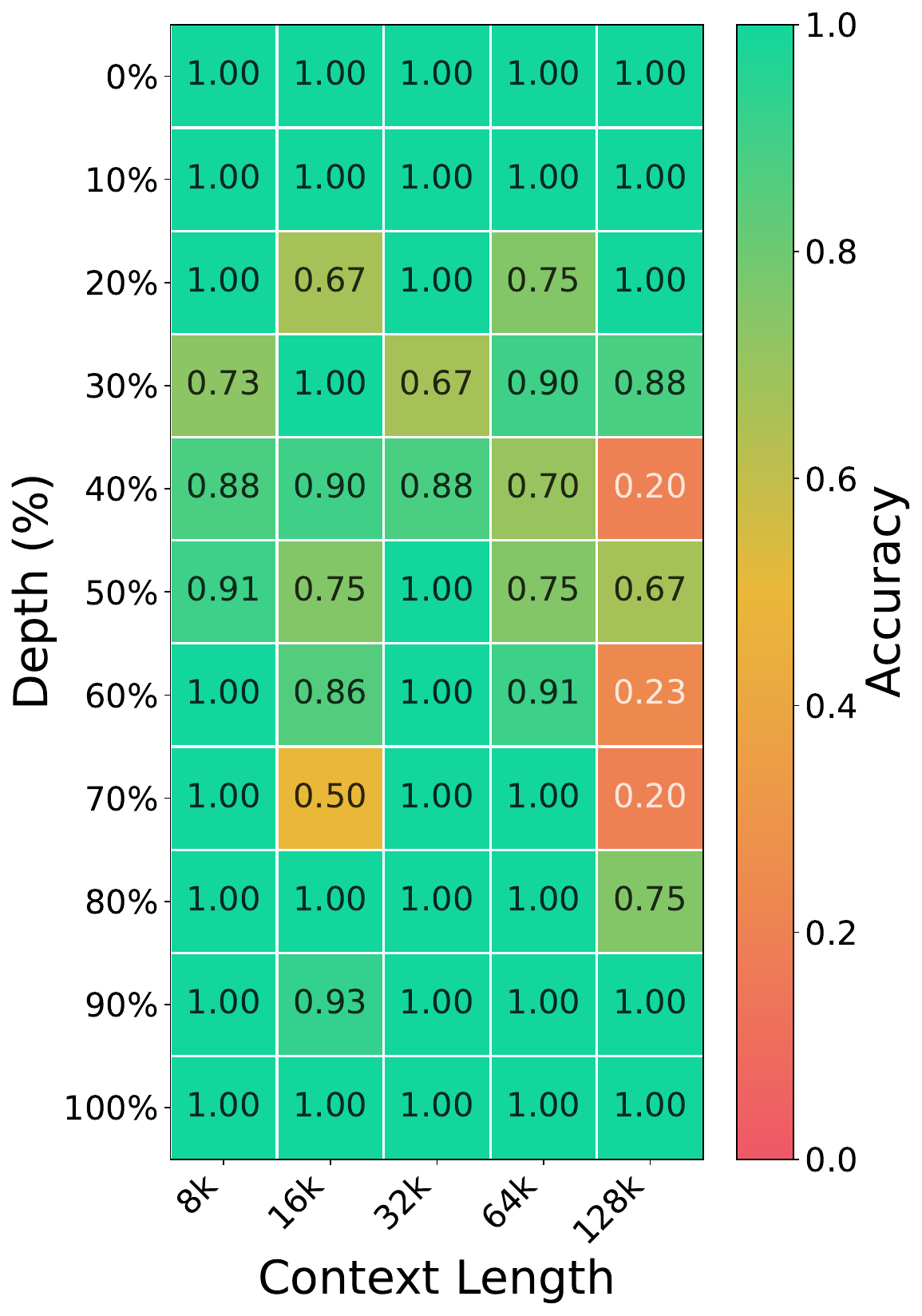}
        \caption{HISA}
        \label{fig:niah_hisa}
    \end{subfigure}
    \hfill
    \begin{subfigure}[t]{0.19\textwidth}
        \centering
        \includegraphics[width=\linewidth]{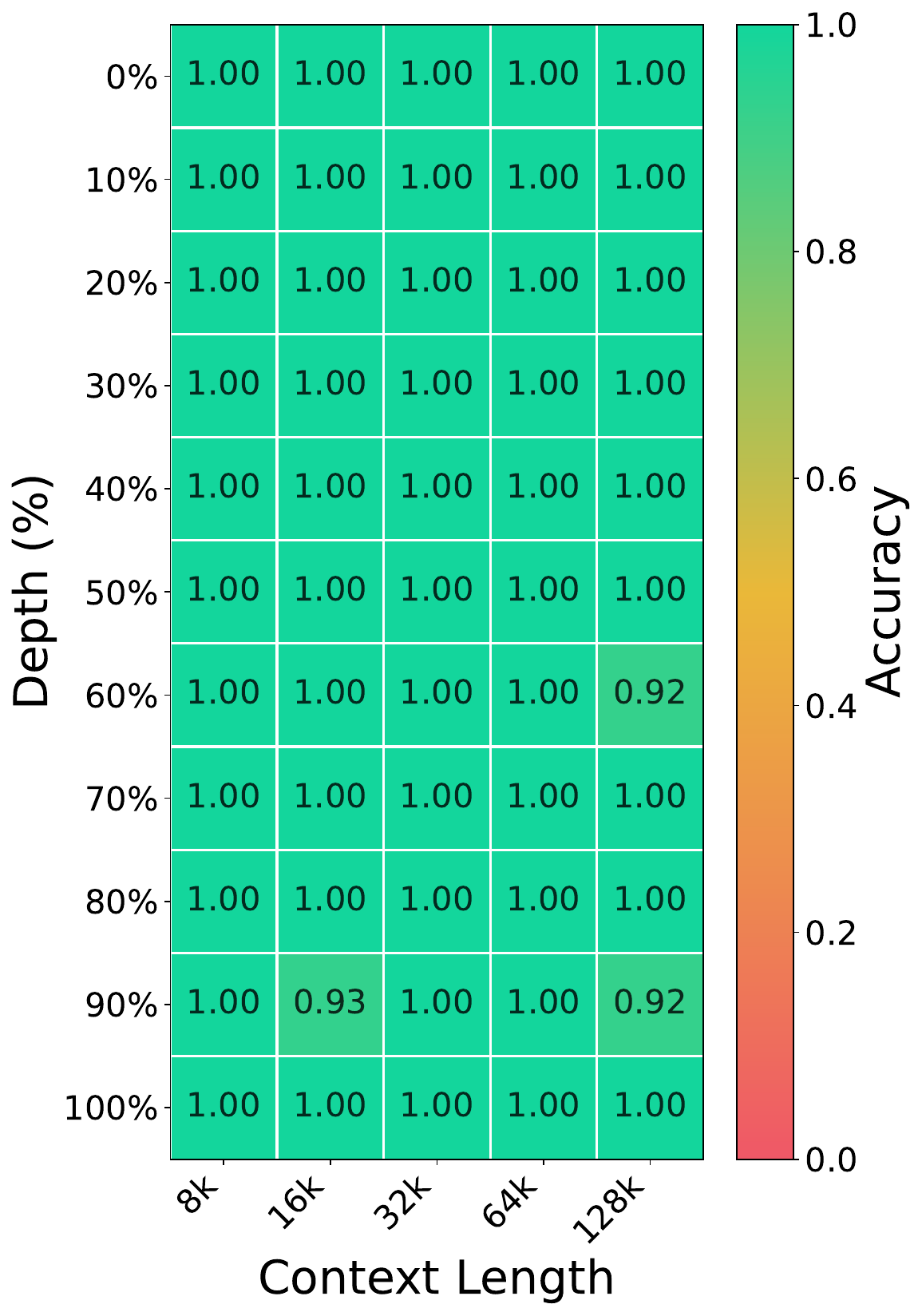}
        \caption{\methodShort}
        \label{fig:niah_misa}
    \end{subfigure}
    \hfill
    \begin{subfigure}[t]{0.19\textwidth}
        \centering
        \includegraphics[width=\linewidth]{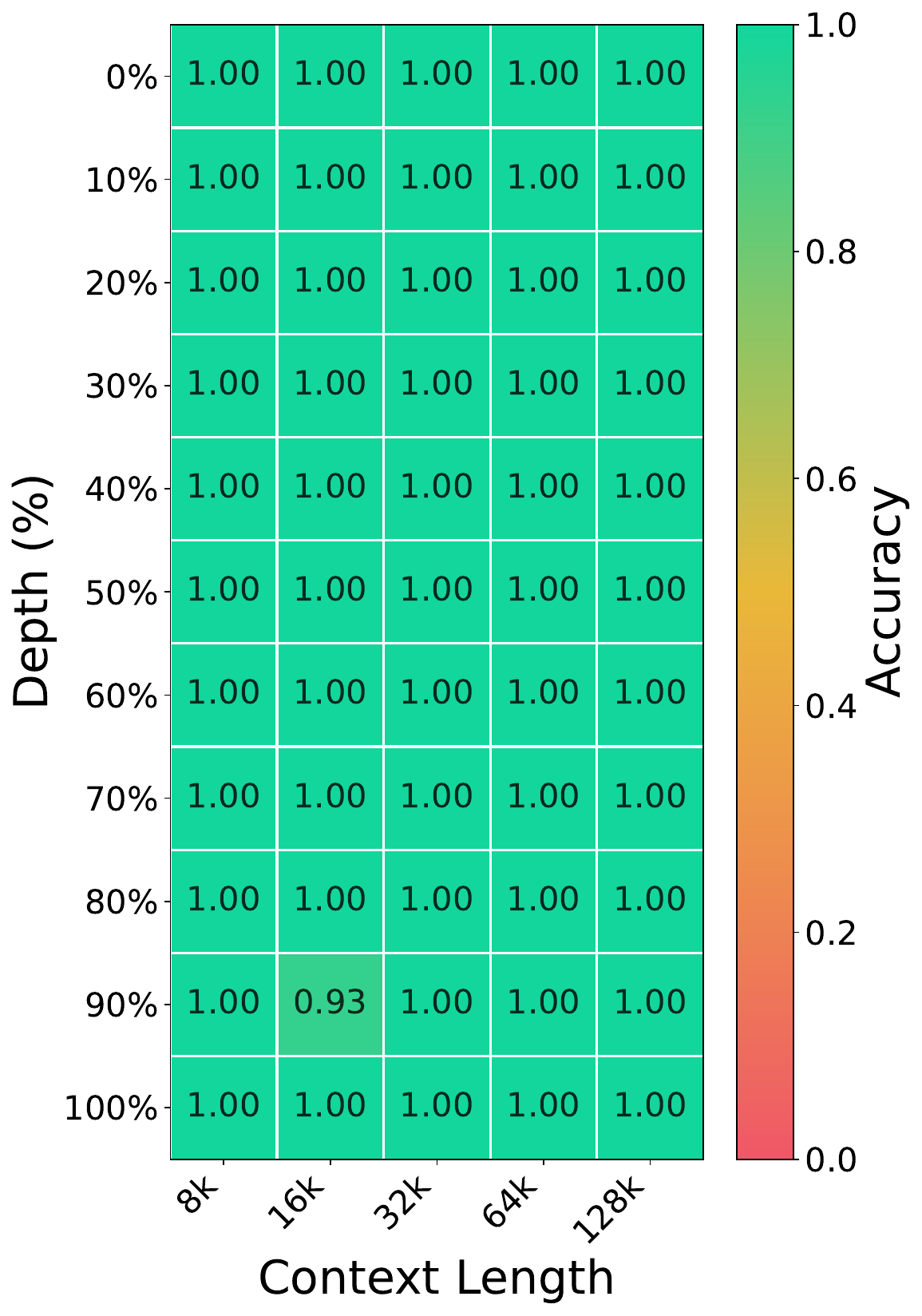}
        \caption{\methodShort$^\dagger$}
        \label{fig:niah_misa_dagger}
    \end{subfigure}
    \hfill
    \caption{Needle-in-a-Haystack retrieval accuracy on DeepSeek-V3.2 up to 128K context. The $x$-axis is context length and the $y$-axis is needle depth (0\%--100\%); greener is better. \methodShort{} and \methodShort$^\dagger$ use only $h = 8$ active indexer heads (vs.\ $H^I = 64$ for the baselines), with $k = 2048$ tokens selected; \methodShort$^\dagger$ additionally uses a coarse candidate set of $k' = 8192$.}
    \label{fig:niah}
    \end{figure}

Figure~\ref{fig:niah} shows the Needle-in-a-Haystack (NIAH) retrieval accuracy of every method on DeepSeek-V3.2 at context lengths up to 128K. For each panel the $x$-axis sweeps the context length and the $y$-axis sweeps the needle depth from $0\%$ (start of the haystack) to $100\%$ (end). \emph{DSA (original)} is the native dense indexer that scores every prefix token with all $H^I = 64$ heads (no token sparsification at the indexer level), and serves as the upper bound. Block-Sparse and HISA use $B = 128$ block partitioning with token budget $k = 2048$; \methodShort{} and \methodShort$^\dagger$ both use only $h = 8$ active heads with $B = 1024$. \methodShort{} selects $k = 2048$ tokens in a single MoE-routed pass, while \methodShort$^\dagger$ first selects a candidate set of size $k' = 8192$ with the same router and then re-ranks those candidates with all $H^I$ DSA heads to keep the final $k = 2048$.

Both \methodShort{} variants reproduce the near-perfect green grid of DSA across the full depth--length plane, in stark contrast to Block-Sparse, whose block-uniform selection leaves visible accuracy holes at intermediate depths once the context exceeds $\sim$32K. HISA closes most of those holes but still shows minor degradations at the deepest needle positions, where its block-level top-$m$ filtering can occasionally drop the block containing the needle. \methodShort$^\dagger$, in particular, is essentially indistinguishable from DSA, confirming that head-level routing combined with a token-level fine pass recovers the dense indexer's retrieval ability while operating with $1/8$ of the per-token head--token products.

\subsection{Indexer kernel speed}
\label{sec:exp_speed}
\begin{figure}[htb]
\centering
\begin{subfigure}[t]{0.45\textwidth}
    \centering
    \includegraphics[width=\linewidth]{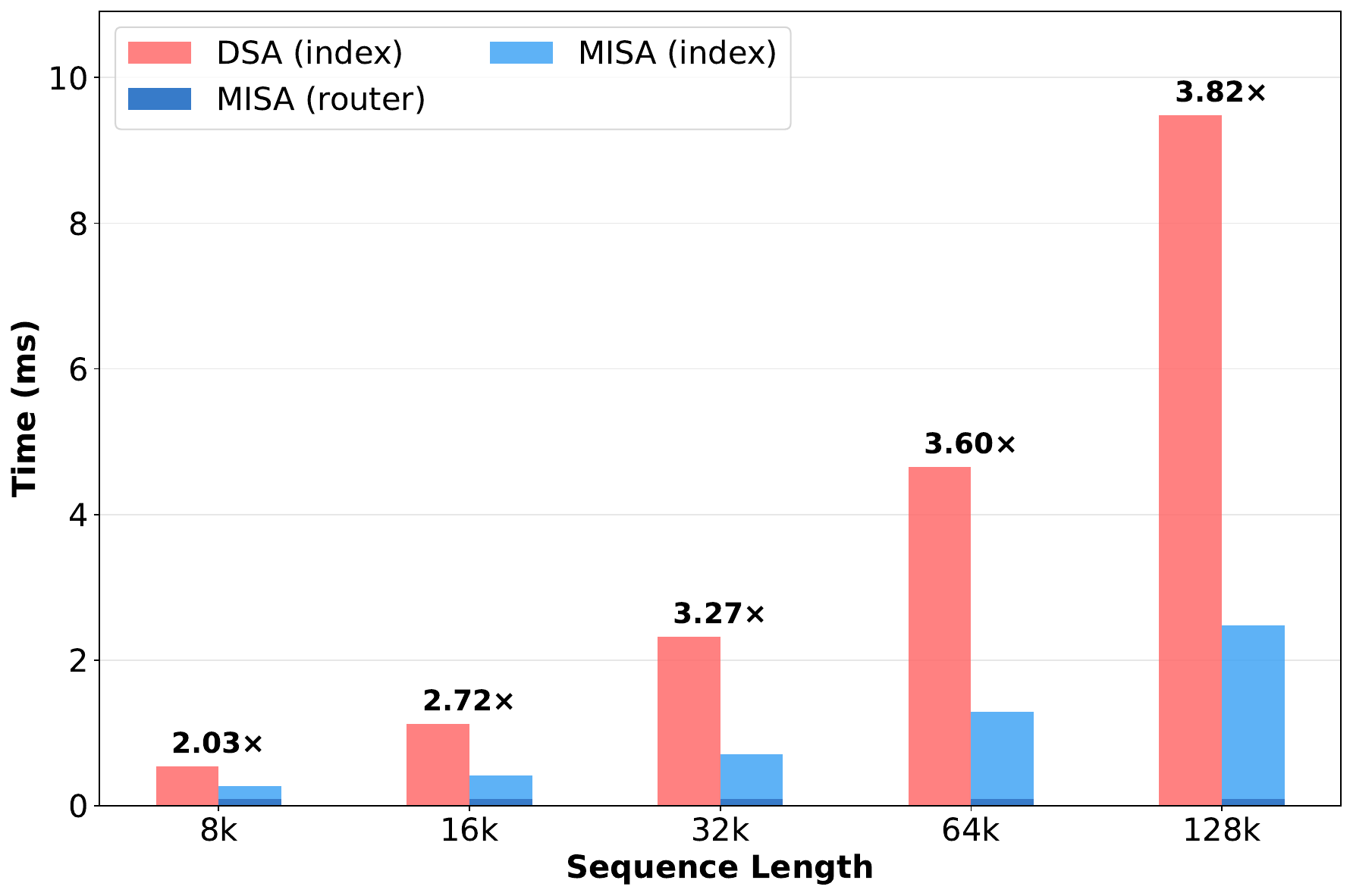}
    \caption{\methodShort}
    \label{fig:speed_1}
\end{subfigure}
\hfill
\begin{subfigure}[t]{0.45\textwidth}
    \centering
    \includegraphics[width=\linewidth]{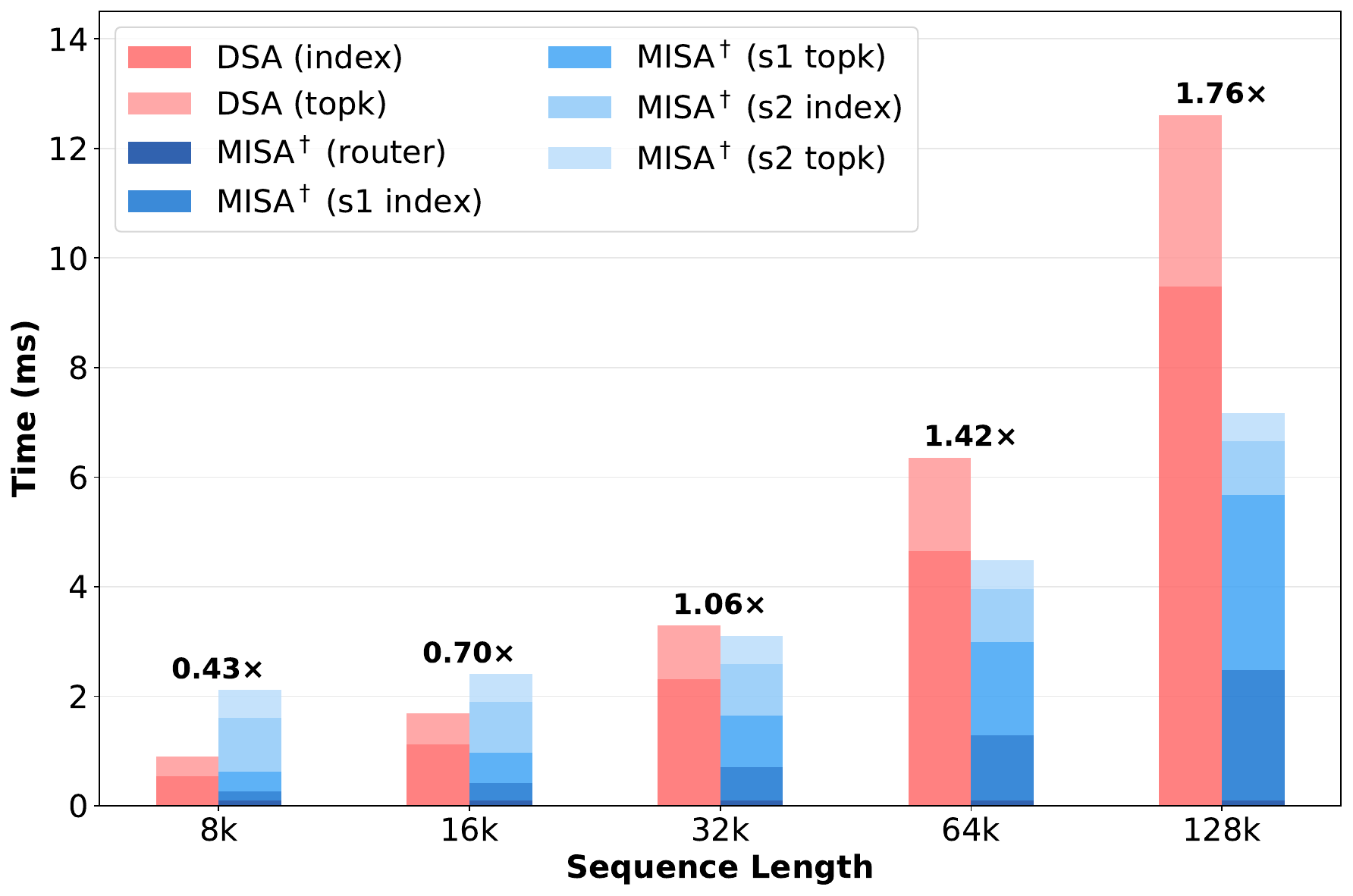}
    \caption{\methodShort$^\dagger$}
    \label{fig:speed_2}
\end{subfigure}
\caption{Indexer-kernel latency on a single NVIDIA H200 GPU for DSA  and \methodShort{}, as a function of prefix length. \methodShort{} uses $h = 8$ active heads with router block size $B = 1024$ and selects $k = 2048$ tokens. \textbf{(a)} \methodShort{} latency. \textbf{(b)} \methodShort$^\dagger$ latency(first stage selects $k^\prime=8912$ tokens). Lower is better.}
\label{fig:speed}
\end{figure}

Figure~\ref{fig:speed} compares the wall-clock latency of the indexer kernel for DSA and \methodShort{} on a single \textbf{NVIDIA H200} GPU, under both the 1-stage (Fig.~\ref{fig:speed_1}) and 2-stage (Fig.~\ref{fig:speed_2},marked as \methodShort$^\dagger$) configurations. DSA scores every prefix token using all $H^I = 64$ indexer heads, resulting in a computational cost that grows as $\mathcal{O}(L H^I)$. In contrast, \methodShort{} activates only $h = 8$ heads per query, scoring the prefix with complexity $\mathcal{O}(L h + M H^I)$, where the second term corresponds to the (negligible) router overhead applied on $M = \lceil L / B \rceil \ll L$ pooled keys. In the 2-stage setting, the complexity becomes $\mathcal{O}(L h + M H^I + k^\prime H^I)$, where $k^\prime$ is fixed to 8192 and accounts for the additional scoring cost in the second stage.

In the 1-stage setting, the \methodShort{} kernel is consistently faster than DSA across the entire sequence-length range. In the 2-stage setting, it also outperforms DSA when the sequence length exceeds 32k. The asymptotic $H^I / h = 8\times$ reduction in head--token products is not fully reflected in wall-clock latency—factors such as memory traffic, load imbalance introduced by the routed expert set, and a small but non-zero router overhead all reduce the theoretical speedup. Nevertheless, our TileLang implementation already achieves approximately a $3.82\times$ end-to-end speedup over DSA’s original indexer kernel for long contexts. This realized
speedup confirms that head-axis routing not only yields the quality benefits reported in Section~\ref{exp}, but also translates into measurable savings on real hardware.

\subsection{Ablation: number  active heads}
\label{sec:exp_nhead}
\begin{figure}[htb]
    \centering
    \begin{subfigure}[t]{0.19\textwidth}
        \centering
        \includegraphics[width=\linewidth]{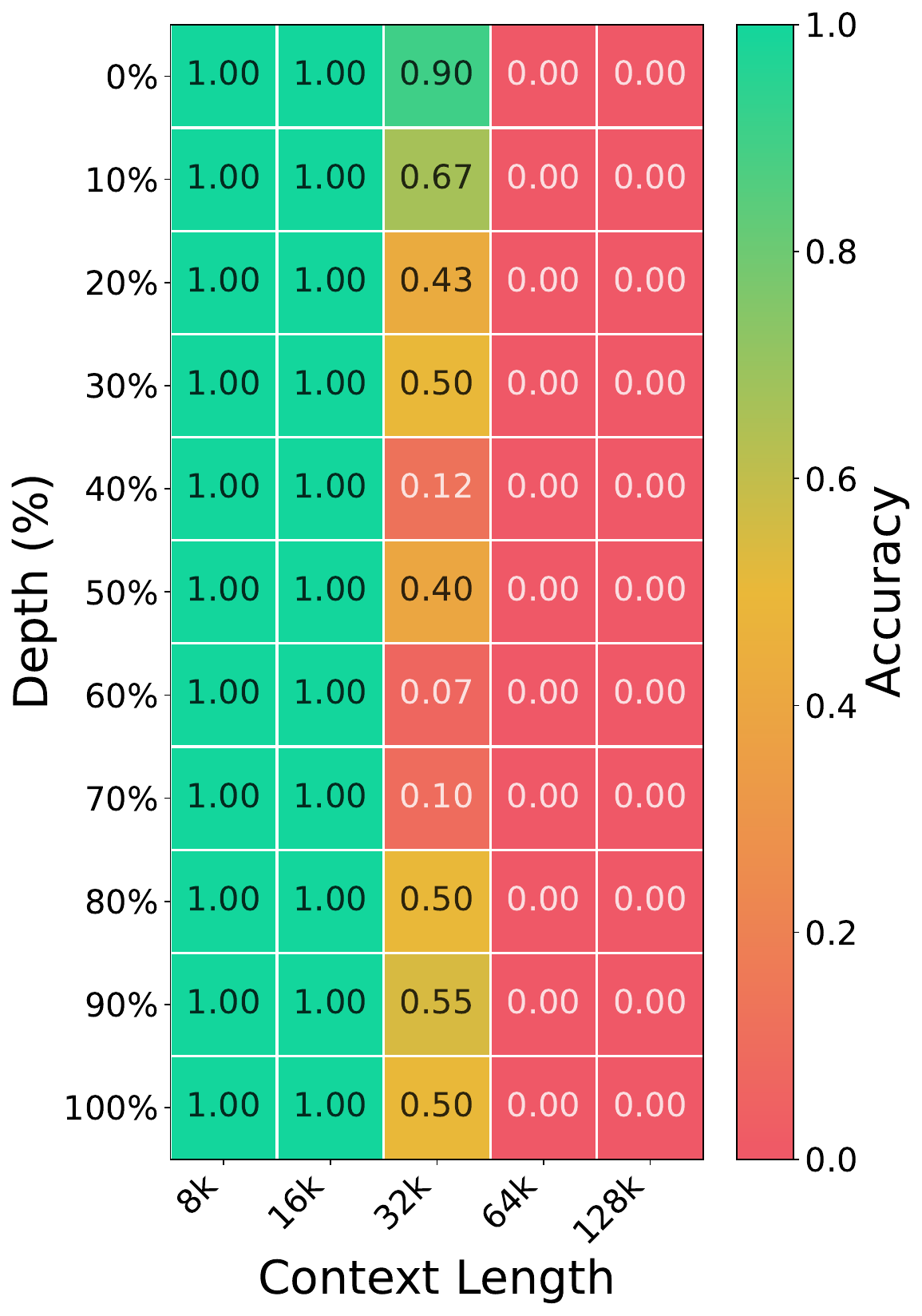}
        \caption{$h = 1$}
        \label{fig:nhead_h1}
    \end{subfigure}
    \hfill
    \begin{subfigure}[t]{0.19\textwidth}
        \centering
        \includegraphics[width=\linewidth]{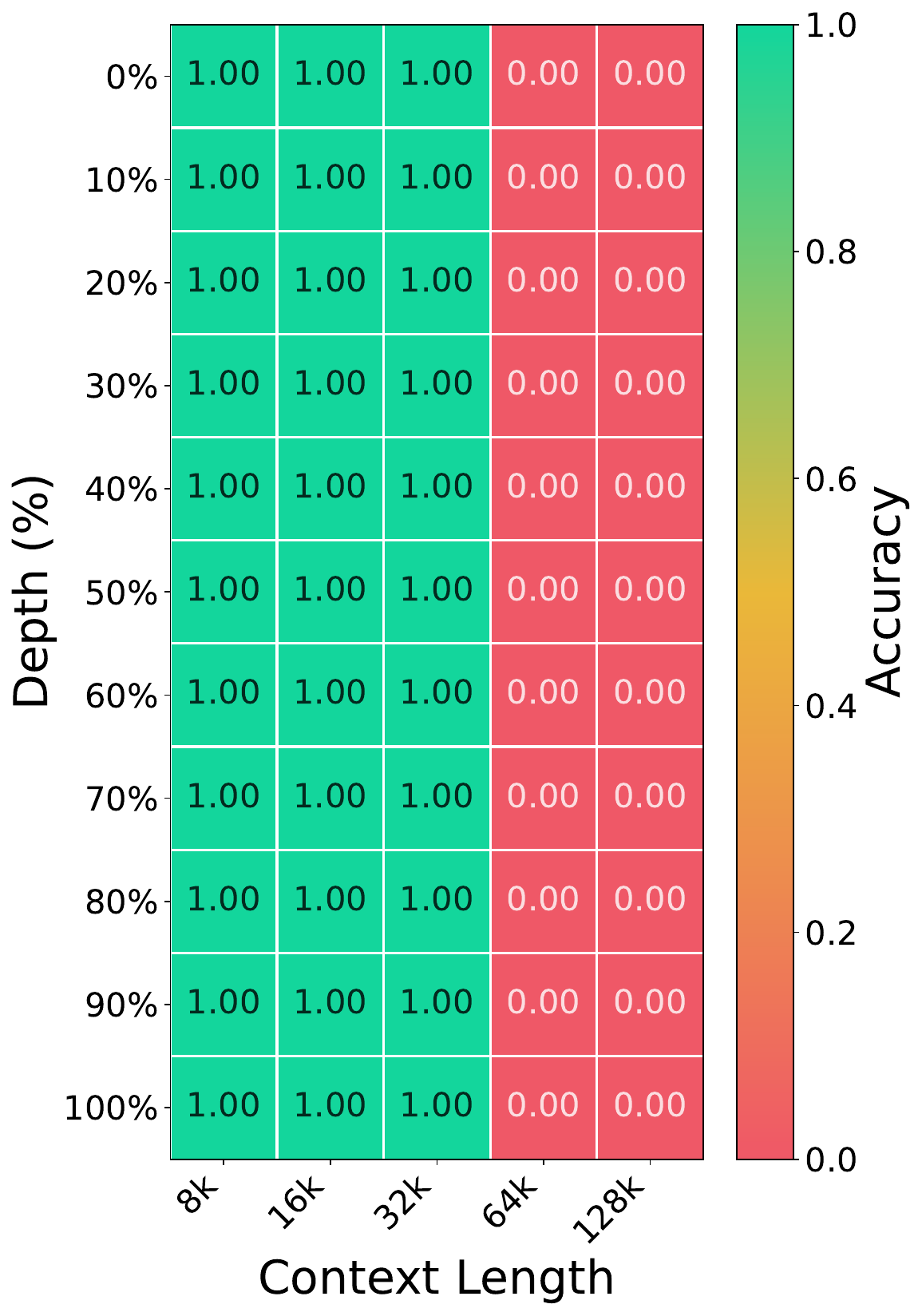}
        \caption{$h = 2$}
        \label{fig:nhead_h2}
    \end{subfigure}
    \hfill
    \begin{subfigure}[t]{0.19\textwidth}
        \centering
        \includegraphics[width=\linewidth]{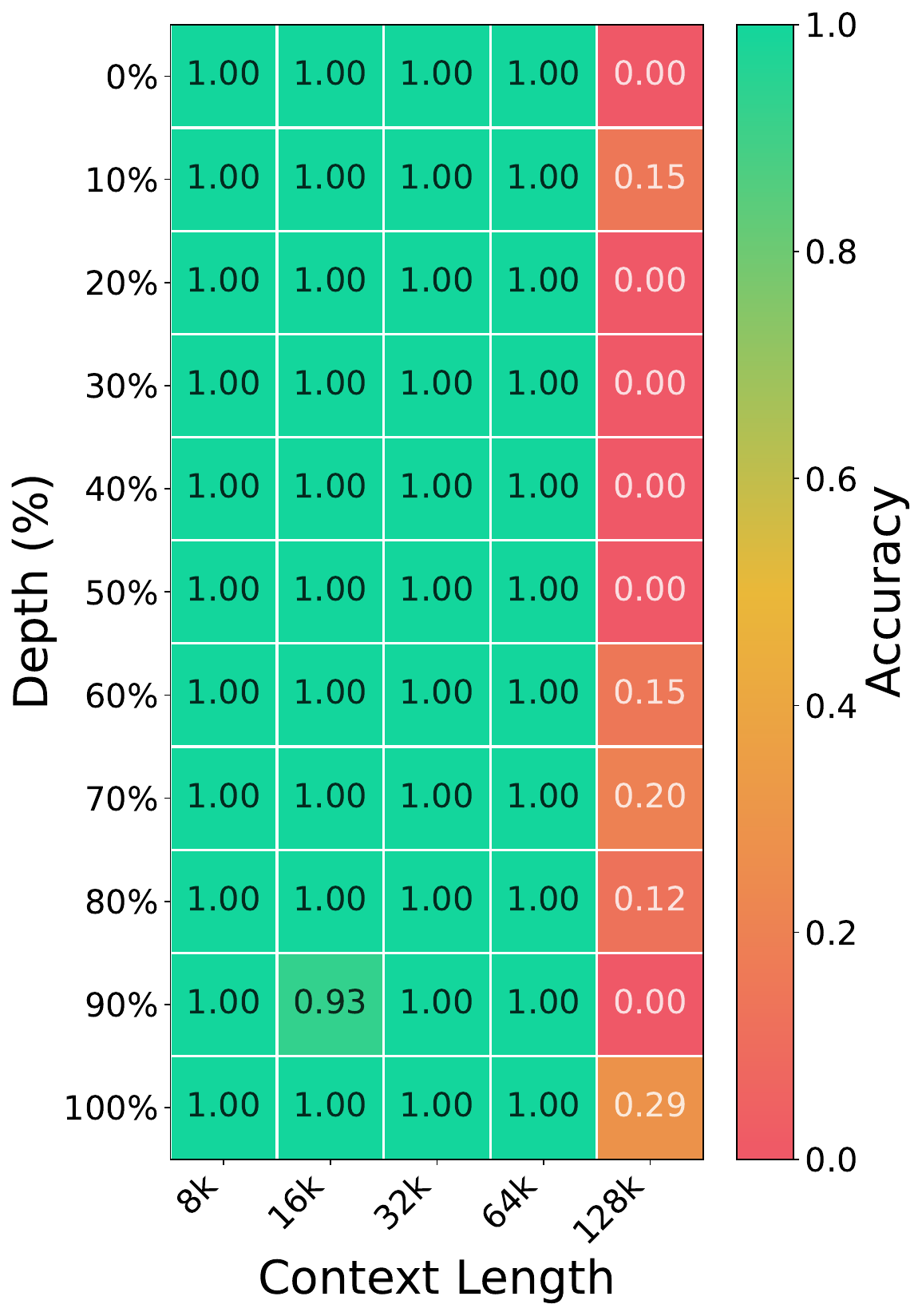}
        \caption{$h = 4$}
        \label{fig:nhead_h4}
    \end{subfigure}
    \hfill
    \begin{subfigure}[t]{0.19\textwidth}
        \centering
        \includegraphics[width=\linewidth]{figures/ablation_head/DeepSeek-v32-mola-2-stage1-h8_niah_single_2_heatmap.pdf}
        \caption{$h = 8$ (default)}
        \label{fig:nhead_h8}
    \end{subfigure}
    \hfill
    \begin{subfigure}[t]{0.19\textwidth}
        \centering
        \includegraphics[width=\linewidth]{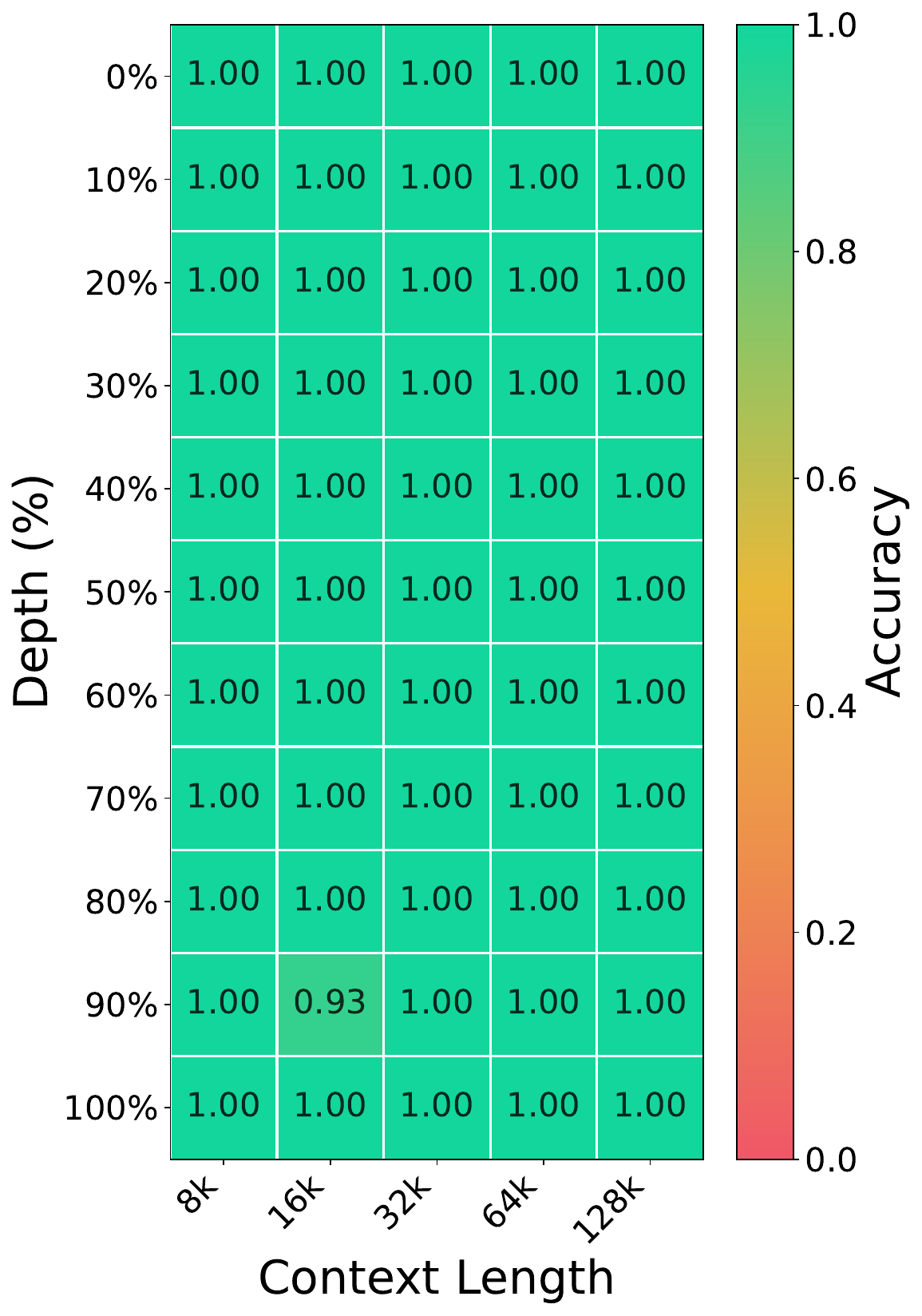}
        \caption{$h = 16$}
        \label{fig:nhead_h16}
    \end{subfigure}
    \caption{Ablation on the number of active heads $h$ used by the \methodShort{} router on DeepSeek-V3.2 ($H^I = 64$, $B = 1024$, $k = 2048$). Each panel is a NIAH retrieval-accuracy heatmap at 128K context, with context length on the $x$-axis and needle depth (0\%--100\%) on the $y$-axis; greener is better.}
    \label{fig:nhead}
    \end{figure}

Figure~\ref{fig:nhead} sweeps the number of active heads $h \in \{1, 2, 4, 8, 16\}$ on DeepSeek-V3.2 with $B = 1024$ and $k = 2048$ fixed, evaluated on NIAH at 128K. As expected, $h = 1$ and $h = 2$ prove too aggressive: one or two routed heads cannot cover the diversity of relevance patterns in DSA's $H^I = 64$-head pool, and the heatmap shows visible accuracy holes. While setting $h = 4$ mitigates most of the aforementioned deficiencies, it continues to demonstrate suboptimal performance, particularly at the maximum evaluated context length of 128K tokens. From $h = 8$ onwards, the heatmap becomes essentially indistinguishable from the dense $64$-head indexer, with $h = 16$ providing no further gain despite using twice the compute of $h = 8$. We therefore default to $h = 8$ as the smallest setting that consistently matches DSA across \emph{every} downstream metric in our experiments. Results for \methodShort$^\dagger$ are provided in Appendix~\ref{app:exp_nhead2}.

\section{Conclusion}
\label{conclusion}
We presented \methodShort{}, which serves as an MoE-style replacement for the DSA indexer. A lightweight block-pooled router selects a query-dependent subset of $h \ll H^I$ active heads, and only those heads run the heavy token-level scan. This single change cuts the dominant per-token cost of fine-grained sparse attention from $\mathcal{O}(H^I L)$ to $\mathcal{O}(h L + H^I M)$ while preserving the full diversity of the indexer pool, because every head remains available---routing simply chooses which ones to consult on each token. A coarse-to-fine extension, \methodShort$^\dagger$, additionally re-ranks an enlarged routed candidate set with the original DSA indexer and recovers the dense top-$k$ almost exactly.

Our experiments support the design on two open-weight long-context models. With $h = 8$ active heads (an $8\times$ head reduction on DeepSeek-V3.2 and a $4\times$ reduction on GLM-5) and \emph{no additional training}, \methodShort{} matches the dense DSA indexer on LongBench within $0.5$ average points, outperforms both Block-Sparse and HISA on average, and retains a fully green Needle-in-a-Haystack heatmap up to 128K context. Our TileLang kernel implementation translates these savings into roughly a $3.82\times$ wall-clock speedup over DSA's original kernel on a single NVIDIA H200 GPU.


\section{Limitation}
\label{limitation}
Several questions regarding \methodShort{} remain unaddressed in this paper: (i) The speed experiments only measure the latency of the TileLang kernel rather than the end-to-end latency of the full model. (ii) Although the MoE-style indexer reduces the computational cost of the DSA stage, it does not reduce the memory access volume to the KV cache in that stage.(iii) All results in this paper are obtained by inserting \methodShort{} into pretrained DSA-based models without finetuning,  jointly training the router with the indexer should further close the residual gap on the few categories where \methodShort{} trails the dense baseline. We will actively explore the aforementioned open questions. Nevertheless, the presented experiments have comprehensively validated the superiority of \methodShort{}. We hope our work will stimulate interest and encourage further exploration within the community.
\newpage

\bibliographystyle{plainnat}
\bibliography{references}

@techreport{openai2026gpt55,
  title       = {Introducing {GPT-5.5}},
  author      = {{OpenAI}},
  year        = {2026},
  institution = {OpenAI},
  note        = {API context window: 1{,}050{,}000 tokens},
  url         = {https://openai.com/index/introducing-gpt-5-5/}
}

@techreport{anthropic2026claudeopus47,
  title       = {Introducing {Claude Opus 4.7}},
  author      = {{Anthropic}},
  year        = {2026},
  institution = {Anthropic},
  note        = {Context window: 1{,}000{,}000 tokens},
  url         = {https://www.anthropic.com/news/claude-opus-4-7}
}

@techreport{google2026gemini3,
  title       = {{Gemini 3}: A New Era of Intelligence},
  author      = {{Google DeepMind}},
  year        = {2026},
  institution = {Google},
  note        = {Context window: 1{,}048{,}576 tokens},
  url         = {https://blog.google/technology/google-deepmind/gemini-3/}
}

@article{yang2025qwen3,
  title   = {{Qwen3} Technical Report},
  author  = {Yang, An and Li, Anfeng and Yang, Baosong and Zhang, Beichen and Hui, Binyuan and Zheng, Bo and Yu, Bowen and Gao, Chang and Huang, Chengen and Lv, Chenxu and others},
  journal = {arXiv preprint arXiv:2505.09388},
  year    = {2025},
  url     = {https://arxiv.org/abs/2505.09388}
}

@article{li2025minimax01,
  title   = {{MiniMax-01}: Scaling Foundation Models with Lightning Attention},
  author  = {Li, Aonian and Gong, Bangwei and Yang, Bo and Shan, Boji and Liu, Chang and Zhu, Cheng and Zhang, Chunhao and Guo, Congchao and Chen, Da and Li, Dong and others},
  journal = {arXiv preprint arXiv:2501.08313},
  year    = {2025},
  url     = {https://arxiv.org/abs/2501.08313}
}

@techreport{moonshot2025kimik2,
  title       = {{Kimi K2}: Open Agentic Intelligence},
  author      = {{Moonshot AI}},
  year        = {2025},
  institution = {Moonshot AI},
  url         = {https://github.com/MoonshotAI/Kimi-K2}
}

@article{deepseekv32,
  title   = {DeepSeek-V3.2: Pushing the Frontier of Open Large Language Models},
  author  = {{DeepSeek-AI}},
  journal = {arXiv preprint arXiv:2512.02556},
  year    = {2025},
  url     = {https://arxiv.org/abs/2512.02556}
}

@article{deepseekv4,
  title   = {{DeepSeek-V4}: Towards Highly Efficient Million-Token Context},
  author  = {{DeepSeek-AI}},
  journal = {Technical Report, DeepSeek},
  year    = {2026},
  url     = {https://huggingface.co/deepseek-ai/DeepSeek-V4-Pro/resolve/main/DeepSeek_V4.pdf}
}

@article{zeng2026glm5,
  title   = {{GLM-5}: from Vibe Coding to Agentic Engineering},
  author  = {Zeng, Aohan and Lv, Xin and Hou, Zhenyu and Du, Zhengxiao and Zheng, Qinkai and Chen, Bin and Yin, Da and Ge, Chendi and Xie, Chengxing and Zhu, Chenzheng and others},
  journal = {arXiv preprint arXiv:2602.15763},
  year    = {2026},
  url     = {https://arxiv.org/abs/2602.15763}
}

@article{bai2026indexcache,
  title   = {{IndexCache}: Accelerating Sparse Attention via Cross-Layer Index Reuse},
  author  = {Bai, Yushi and Dong, Qian and Jiang, Ting and Lv, Xin and Du, Zhengxiao and Zeng, Aohan and Tang, Jie and Li, Juanzi},
  journal = {arXiv preprint arXiv:2603.12201},
  year    = {2026},
  url     = {https://arxiv.org/abs/2603.12201}
}

@article{enzhe2025moba,
  title={{MoBA}: Mixture of Block Attention for Long-Context {LLM}s},
  author={Lu, Enzhe and Jiang, Zhejun and Liu, Jingyuan and Du, Yulun and Jiang, Tao and Hong, Chao and Liu, Shaowei and He, Weiran and Yuan, Enming and Wang, Yuzhi and Huang, Zhiqi and Yuan, Huan and Xu, Suting and Xu, Xinran and Lai, Guokun and Chen, Yanru and Zheng, Huabin and Yan, Junjie and Su, Jianlin and Wu, Yuxin and Zhang, Neo Y. and Yang, Zhilin and Zhou, Xinyu and Zhang, Mingxing and Qiu, Jiezhong},
  journal={arXiv preprint arXiv:2502.13189},
  year={2025},
  url={https://www.arxiv.org/abs/2502.13189}
}

@article{yufei2026hisa,
  title={{HISA}: Efficient Hierarchical Indexing for Fine-Grained Sparse Attention},
  author={Xu, Yufei and Meng, Fanxu and Jiang, Fan and Wang, Yuxuan and Zhou, Ruijie and Wang, Zhaohui and Wu, Jiexi and Pan, Zhixin and Tang, Xiaojuan and Pei, Wenjie and Liu, Tongxuan and Yin, Di and Sun, Xing and Zhang, Muhan},
  journal={arXiv preprint arXiv:2603.28458},
  year={2026},
  url={https://www.arxiv.org/abs/2603.28458v3}
}

@article{child2019sparse,
  title={Generating Long Sequences with Sparse Transformers},
  author={Child, Rewon and Gray, Scott and Radford, Alec and Sutskever, Ilya},
  journal={arXiv preprint arXiv:1904.10509},
  year={2019},
  url={https://arxiv.org/abs/1904.10509}
}

@article{beltagy2020longformer,
  title={Longformer: The Long-Document Transformer},
  author={Beltagy, Iz and Peters, Matthew E. and Cohan, Arman},
  journal={arXiv preprint arXiv:2004.05150},
  year={2020},
  url={https://arxiv.org/abs/2004.05150}
}

@inproceedings{zaheer2020bigbird,
  title={Big Bird: Transformers for Longer Sequences},
  author={Zaheer, Manzil and Guruganesh, Guru and Dubey, Avinava and Ainslie, Joshua and Alberti, Chris and Onta{\~n}{\'o}n, Santiago and Pham, Philip and Ravula, Anirudh and Wang, Qifan and Yang, Li and Ahmed, Amr},
  booktitle={Advances in Neural Information Processing Systems},
  year={2020},
  url={https://arxiv.org/abs/2007.14062}
}

@inproceedings{xiao2024streamingllm,
  title={Efficient Streaming Language Models with Attention Sinks},
  author={Xiao, Guangxuan and Tian, Yuandong and Chen, Beidi and Han, Song and Lewis, Mike},
  booktitle={International Conference on Learning Representations},
  year={2024},
  url={https://arxiv.org/abs/2309.17453}
}

@inproceedings{zhang2023h2o,
  title={{H$_2$O}: Heavy-Hitter Oracle for Efficient Generative Inference of Large Language Models},
  author={Zhang, Zhenyu and Sheng, Ying and Zhou, Tianyi and Chen, Tianlong and Zheng, Lianmin and Cai, Ruisi and Song, Zhao and Tian, Yuandong and R{\'e}, Christopher and Barrett, Clark and Wang, Zhangyang and Chen, Beidi},
  booktitle={Advances in Neural Information Processing Systems},
  year={2023},
  url={https://arxiv.org/abs/2306.14048}
}

@article{li2024snapkv,
  title={{SnapKV}: {LLM} Knows What You are Looking for Before Generation},
  author={Li, Yuhong and Huang, Yingbing and Yang, Bowen and Venkitesh, Bharat and Locatelli, Acyr and Ye, Hanchen and Cai, Tianle and Lewis, Patrick and Chen, Deming},
  journal={arXiv preprint arXiv:2404.14469},
  year={2024},
  url={https://arxiv.org/abs/2404.14469}
}

@inproceedings{tang2024quest,
  title={{Quest}: Query-Aware Sparsity for Efficient Long-Context {LLM} Inference},
  author={Tang, Jiaming and Zhao, Yilong and Zhu, Kan and Xiao, Guangxuan and Kasikci, Baris and Han, Song},
  booktitle={International Conference on Machine Learning},
  year={2024},
  url={https://arxiv.org/abs/2406.10774}
}

@article{xiao2024infllm,
  title={{InfLLM}: Training-Free Long-Context Extrapolation for {LLM}s with an Efficient Context Memory},
  author={Xiao, Chaojun and Zhang, Pengle and Han, Xu and Xiao, Guangxuan and Lin, Yankai and Zhang, Zhengyan and Liu, Zhiyuan and Han, Song and Sun, Maosong},
  journal={arXiv preprint arXiv:2402.04617},
  year={2024},
  url={https://arxiv.org/abs/2402.04617}
}

@article{chen2024magicpig,
  title={{MagicPIG}: {LSH} Sampling for Efficient {LLM} Generation},
  author={Chen, Zhuoming and Sadhukhan, Ranajoy and Ye, Zihao and Zhou, Yang and Zhang, Jianyu and Nolte, Niklas and Tian, Yuandong and Douze, Matthijs and Bottou, Leon and Jia, Zhihao and Chen, Beidi},
  journal={arXiv preprint arXiv:2410.16179},
  year={2024},
  url={https://arxiv.org/abs/2410.16179}
}

@article{gao2024seer,
  title={{SeerAttention}: Learning Intrinsic Sparse Attention in Your {LLM}s},
  author={Gao, Yizhao and Zeng, Zhichen and Du, Dayou and Cao, Shijie and So, Hayden Kwok-Hay and Cao, Ting and Yang, Fan and Yang, Mao},
  journal={arXiv preprint arXiv:2410.13276},
  year={2024},
  url={https://arxiv.org/abs/2410.13276}
}

@article{yuan2025nsa,
  title={Native Sparse Attention: Hardware-Aligned and Natively Trainable Sparse Attention},
  author={Yuan, Jingyang and Gao, Huazuo and Dai, Damai and Luo, Junyu and Zhao, Liang and Zhang, Zhengyan and Xie, Zhenda and Wei, Y.X. and Wang, Lean and Xiao, Zhiping and Wang, Yuqing and Ruan, Chong and Zhang, Ming and Liang, Wenfeng and Wang, Wangding},
  journal={arXiv preprint arXiv:2502.11089},
  year={2025},
  url={https://arxiv.org/abs/2502.11089}
}

@inproceedings{shazeer2017moe,
  title={Outrageously Large Neural Networks: The Sparsely-Gated Mixture-of-Experts Layer},
  author={Shazeer, Noam and Mirhoseini, Azalia and Maziarz, Krzysztof and Davis, Andy and Le, Quoc and Hinton, Geoffrey and Dean, Jeff},
  booktitle={International Conference on Learning Representations},
  year={2017},
  url={https://arxiv.org/abs/1701.06538}
}

@article{lepikhin2020gshard,
  title={{GShard}: Scaling Giant Models with Conditional Computation and Automatic Sharding},
  author={Lepikhin, Dmitry and Lee, HyoukJoong and Xu, Yuanzhong and Chen, Dehao and Firat, Orhan and Huang, Yanping and Krikun, Maxim and Shazeer, Noam and Chen, Zhifeng},
  journal={arXiv preprint arXiv:2006.16668},
  year={2020},
  url={https://arxiv.org/abs/2006.16668}
}

@article{fedus2022switch,
  title={Switch Transformers: Scaling to Trillion Parameter Models with Simple and Efficient Sparsity},
  author={Fedus, William and Zoph, Barret and Shazeer, Noam},
  journal={Journal of Machine Learning Research},
  volume={23},
  number={120},
  pages={1--39},
  year={2022},
  url={https://arxiv.org/abs/2101.03961}
}

@article{jiang2024mixtral,
  title={Mixtral of Experts},
  author={Jiang, Albert Q. and Sablayrolles, Alexandre and Roux, Antoine and Mensch, Arthur and Savary, Blanche and Bamford, Chris and Chaplot, Devendra Singh and de las Casas, Diego and Hanna, Emma Bou and Bressand, Florian and others},
  journal={arXiv preprint arXiv:2401.04088},
  year={2024},
  url={https://arxiv.org/abs/2401.04088}
}

@article{dai2024deepseekmoe,
  title={{DeepSeekMoE}: Towards Ultimate Expert Specialization in Mixture-of-Experts Language Models},
  author={Dai, Damai and Deng, Chengqi and Zhao, Chenggang and Xu, R.X. and Gao, Huazuo and Chen, Deli and Li, Jiashi and Zeng, Wangding and Yu, Xingkai and Wu, Y. and others},
  journal={arXiv preprint arXiv:2401.06066},
  year={2024},
  url={https://arxiv.org/abs/2401.06066}
}

@inproceedings{zhang2022moa,
  title={Mixture of Attention Heads: Selecting Attention Heads Per Token},
  author={Zhang, Xiaofeng and Shen, Yikang and Huang, Zeyu and Zhou, Jie and Rong, Wenge and Xiong, Zhang},
  booktitle={Conference on Empirical Methods in Natural Language Processing},
  year={2022},
  url={https://arxiv.org/abs/2210.05144}
}

@article{jin2024moh,
  title={{MoH}: Multi-Head Attention as Mixture-of-Head Attention},
  author={Jin, Peng and Zhu, Bo and Yuan, Li and Yan, Shuicheng},
  journal={arXiv preprint arXiv:2410.11842},
  year={2024},
  url={https://arxiv.org/abs/2410.11842}
}
\newpage


\appendix
\section{Per-layer agreement with the DSA top-\texorpdfstring{$k$}{k}}
\label{app:exp_iou}
As an intrinsic check that head-level routing does not distort the selected token set, we measure the per-layer Intersection-over-Union (IoU) between each method's selected tokens and the DSA top-$2048$ \emph{reference} set. Concretely, on the long subset of \texttt{LSHT} (10 examples, all $T$ layers, on DeepSeek-V3.2), we run DSA with all $H^I = 64$ indexer heads to obtain the reference token set $\mathcal{T}_t^{\mathrm{DSA}}$, and for every other method we report
\begin{equation}
\mathrm{IoU}_t^{(\ell)} = \frac{\left|\mathcal{T}_t \cap \mathcal{T}_t^{\mathrm{DSA}}\right|}{\left|\mathcal{T}_t \cup \mathcal{T}_t^{\mathrm{DSA}}\right|}, \qquad \ell = 1,\ldots,T.
\label{eq:iou}
\end{equation}
The IoU becomes meaningful only once the prefix length meets or exceeds the token budget; therefore, we initiate the curves at position $t = 2048$. Both \methodShort{} and HISA conclude at the identical final budget of $k = 2048$. To facilitate a fair comparison under an equivalent computational budget, we evaluate \methodShort{} in a two-stage configuration, denoted as \methodShort$^\dagger$: the shared router first generates a candidate set of size $k^\prime = 8192$, which is subsequently re-scored by all $H^I = 64$ DSA heads to select the final $k = 2048$ tokens.

\begin{figure}[htb]
\centering
\begin{subfigure}[t]{0.45\textwidth}
    \centering
    \includegraphics[width=\linewidth]{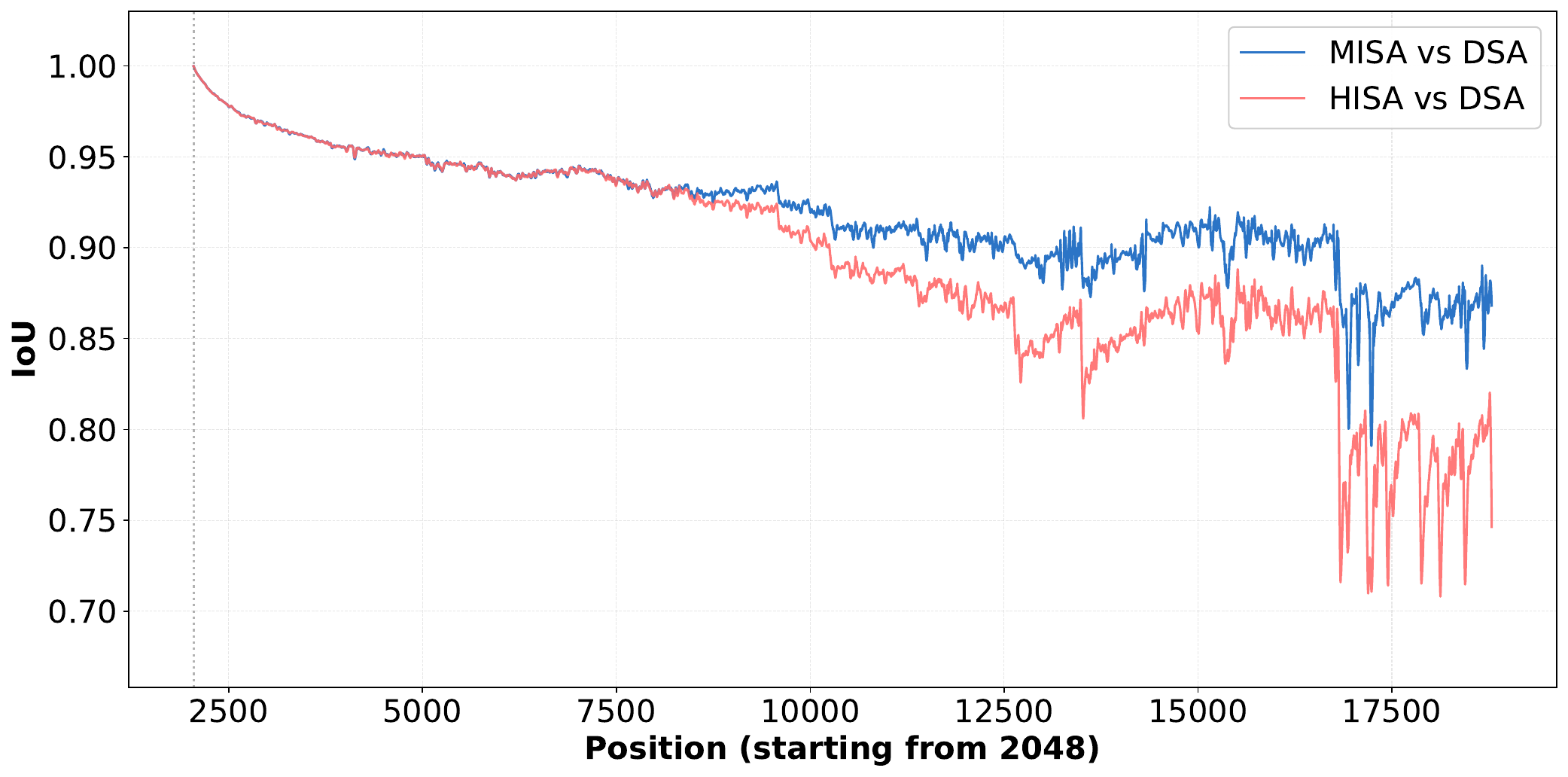}
    \caption{IoU vs.\ token position}
    \label{fig:iou_len}
\end{subfigure}
\hfill
\begin{subfigure}[t]{0.45\textwidth}
    \centering
    \includegraphics[width=\linewidth]{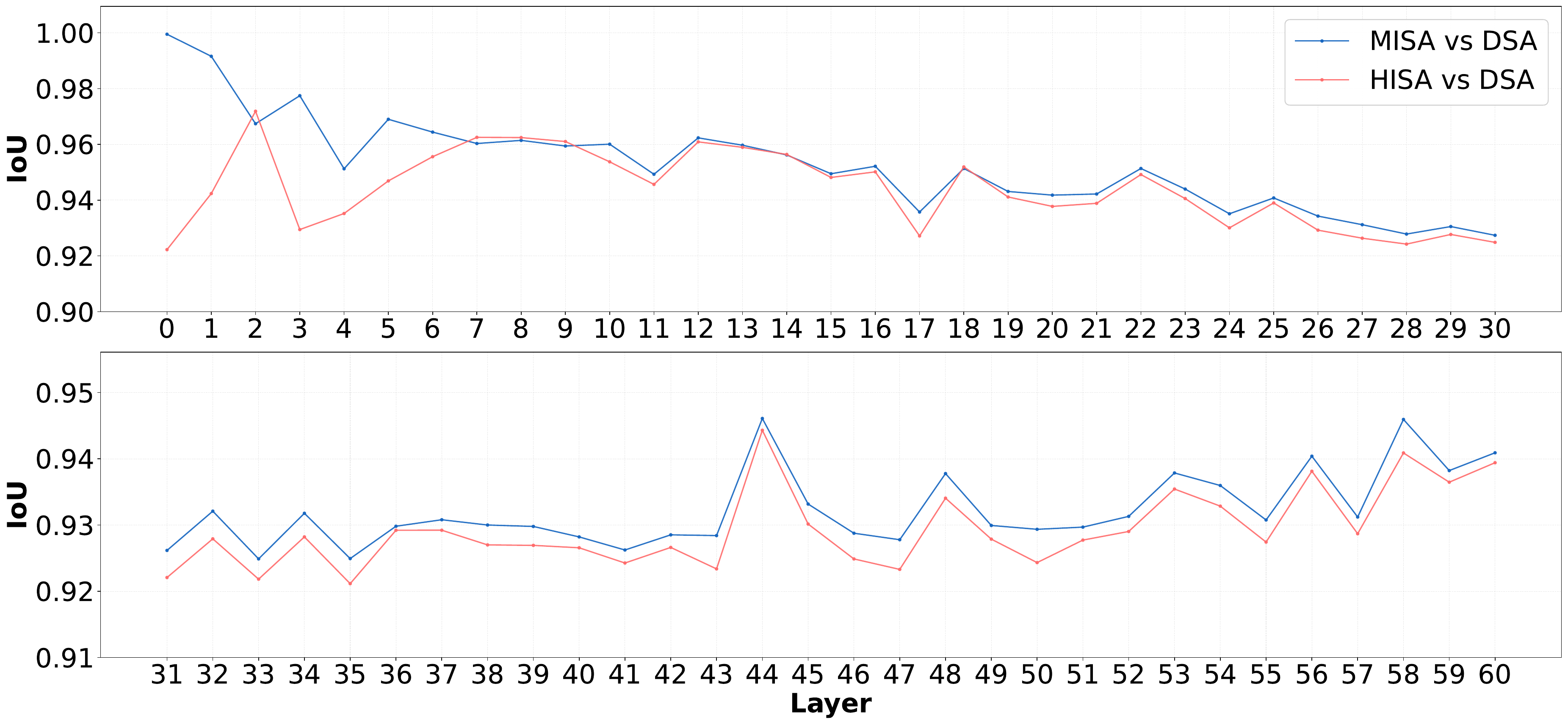}
    \caption{IoU vs.\ layer index}
    \label{fig:iou_layer}
\end{subfigure}
\caption{Per-layer Intersection-over-Union between the indexer-selected token set and the DSA top-$2048$ golden set on \texttt{LSHT}. Curves are plotted for HISA and \methodShort{}-stage-2 (the fine pass of \methodShort$^\dagger$, which re-ranks the $k^\prime = 8192$ candidates with all $H^I$ heads to keep $k = 2048$). \textbf{(a)} IoU as a function of token position, averaged across $T$ layers; the position axis starts at $2048$ since IoU is only well-defined once the prefix has at least $k$ tokens. \textbf{(b)} IoU as a function of layer index.}
\label{fig:iou}
\end{figure}

Figure~\ref{fig:iou} summarizes the results. In each panel, the Intersection-over-Union (IoU) between the token sets selected by \methodShort$^\dagger$ and the DSA baseline is represented by the blue curve, whereas the red curve corresponds to the IoU between HISA and the DSA baseline. Figure~\ref{fig:iou_len} shows how the average IoU varies with sequence length, and Figure~\ref{fig:iou_layer} compares the IoU layer by layer. \methodShort$^\dagger$ consistently outperforms HISA across sequence lengths and at every layer, retaining more than $92\%$ of the tokens selected by the DSA indexer in each layer. This high degree of intrinsic alignment with the DSA top-$k$ set provides the underlying explanation for the empirical result: it is the mechanism that enables \methodShort$^\dagger$ to match the downstream performance of the full DSA indexer on benchmarks such as NIAH and LongBench (see Sections~\ref{sec:exp_niah} and~\ref{sec:exp_longbench}), despite not undergoing any task-specific retraining.

\newpage
\section{Ablation: active heads' number of \texorpdfstring{\methodShort$^\dagger$}{MISA-dagger}}
\label{app:exp_nhead2}
\begin{figure}[htp]
    \centering
    \begin{subfigure}[t]{0.19\textwidth}
        \centering
        \includegraphics[width=\linewidth]{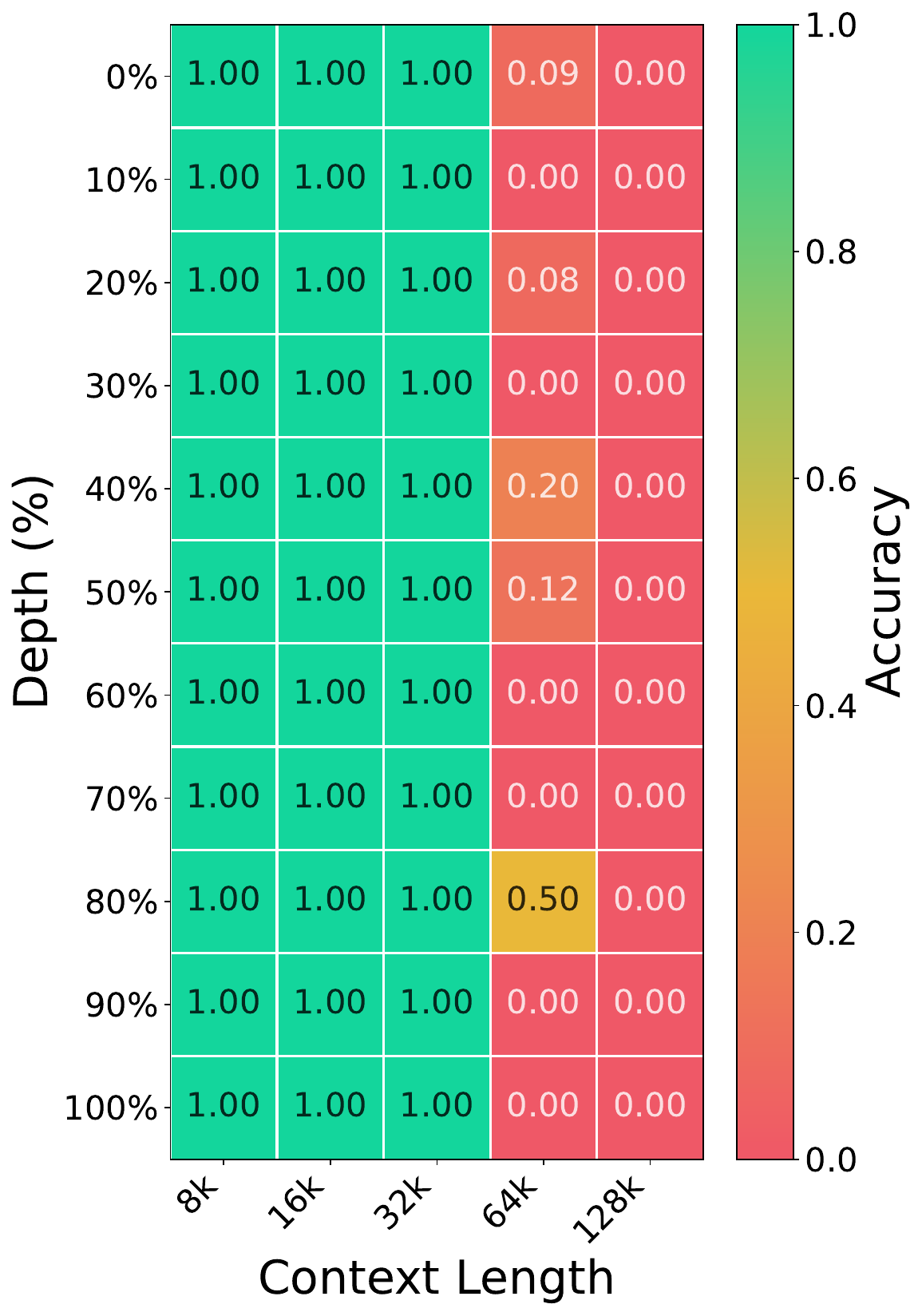}
        \caption{$h = 1$}
        \label{fig:nhead_stage2_h1}
    \end{subfigure}
    \hfill
    \begin{subfigure}[t]{0.19\textwidth}
        \centering
        \includegraphics[width=\linewidth]{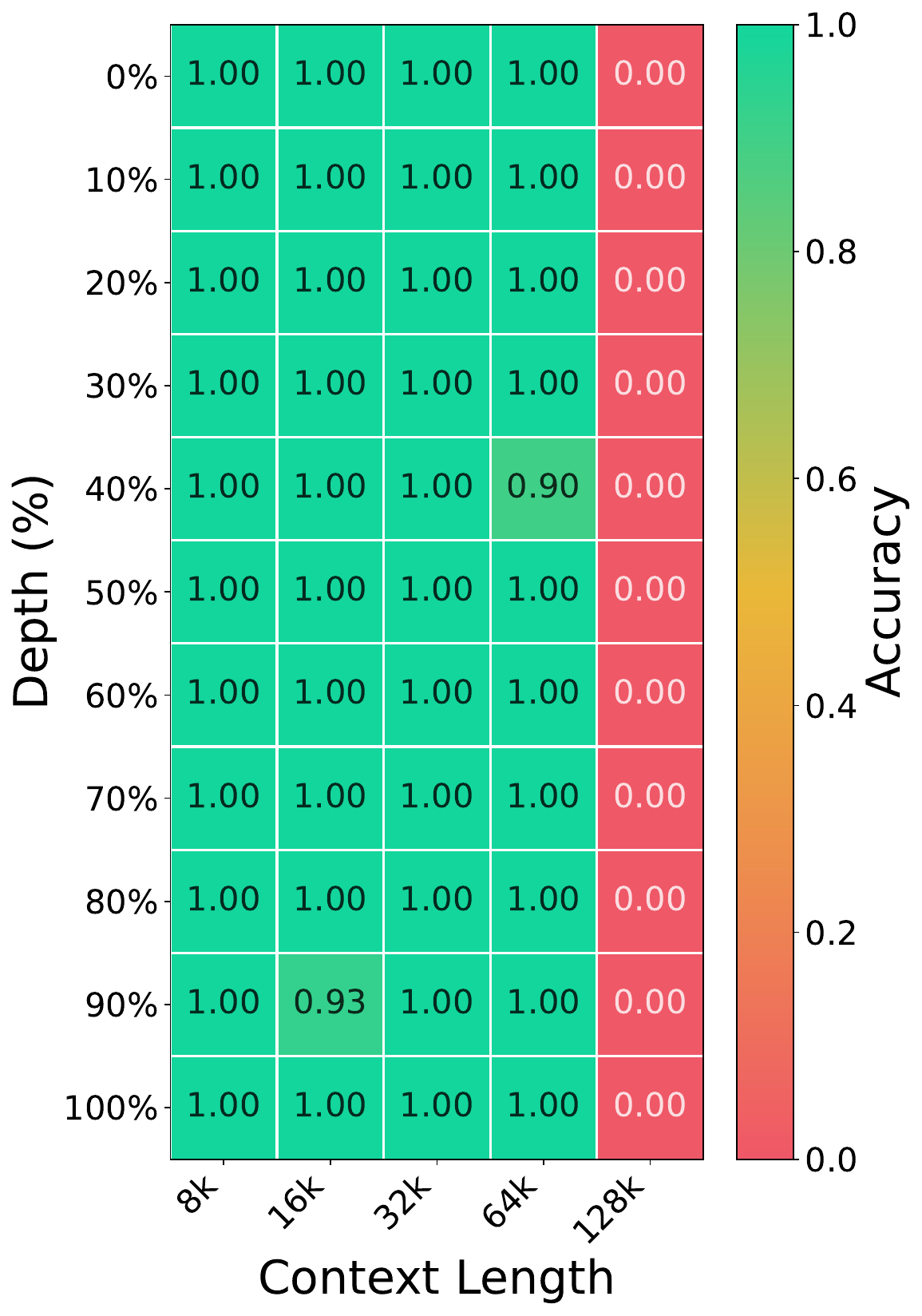}
        \caption{$h = 2$}
        \label{fig:nhead_stage2_h2}
    \end{subfigure}
    \hfill
    \begin{subfigure}[t]{0.19\textwidth}
        \centering
        \includegraphics[width=\linewidth]{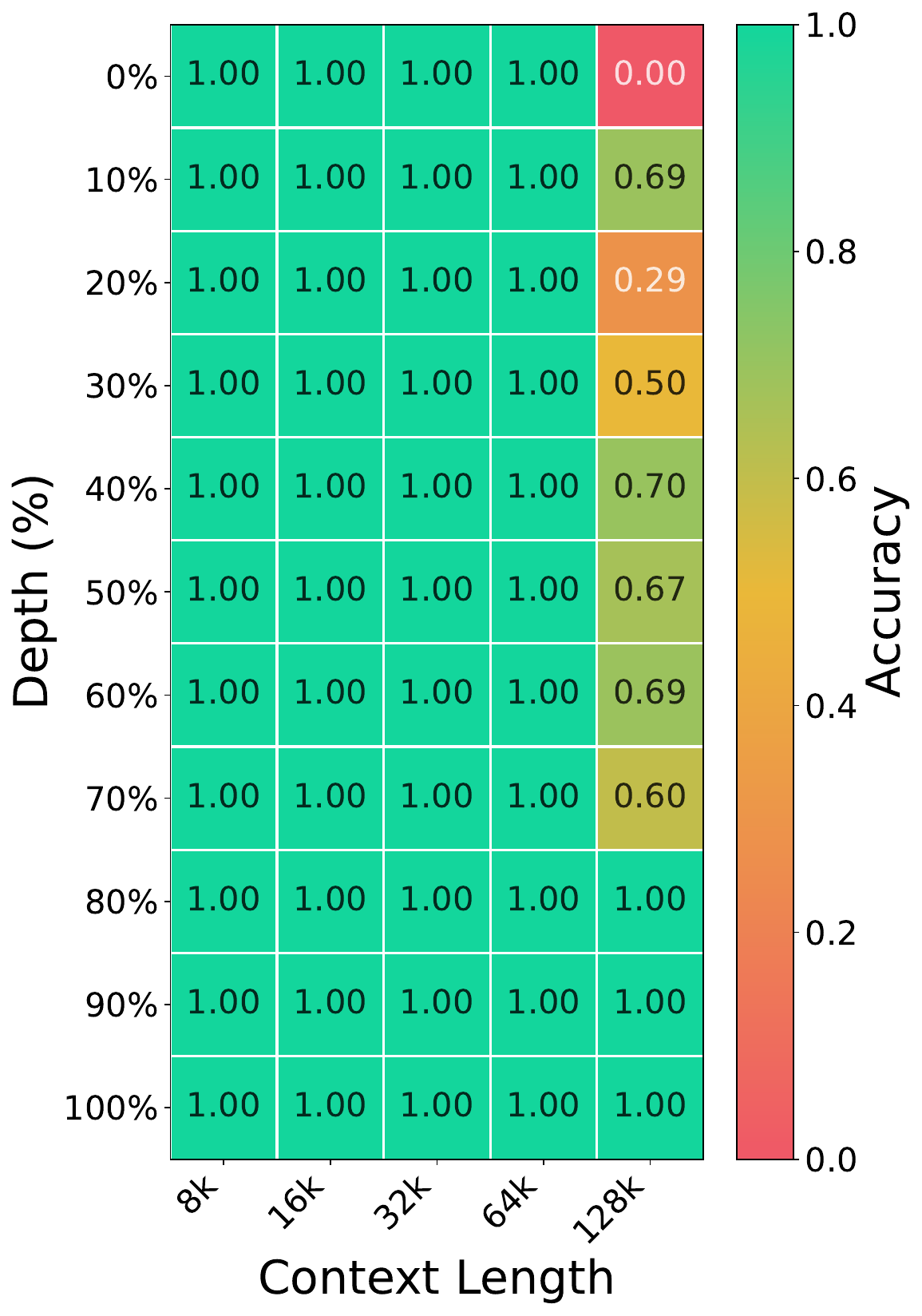}
        \caption{$h = 4$}
        \label{fig:nhead_stage2_h4}
    \end{subfigure}
    \hfill
    \begin{subfigure}[t]{0.19\textwidth}
        \centering
        \includegraphics[width=\linewidth]{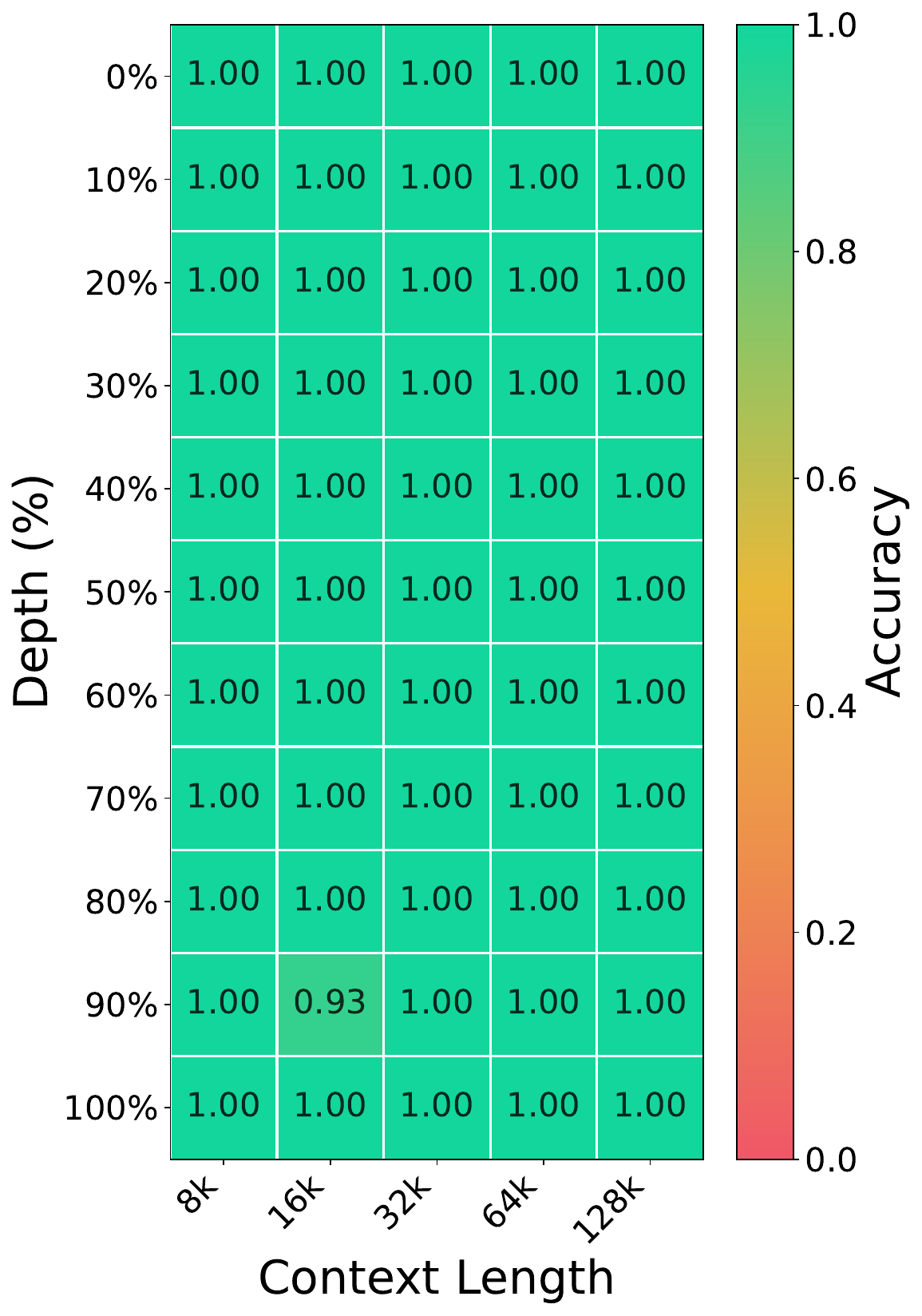}
        \caption{$h = 8$ (default)}
        \label{fig:nhead_stage2_h8}
    \end{subfigure}
    \hfill
    \begin{subfigure}[t]{0.19\textwidth}
        \centering
        \includegraphics[width=\linewidth]{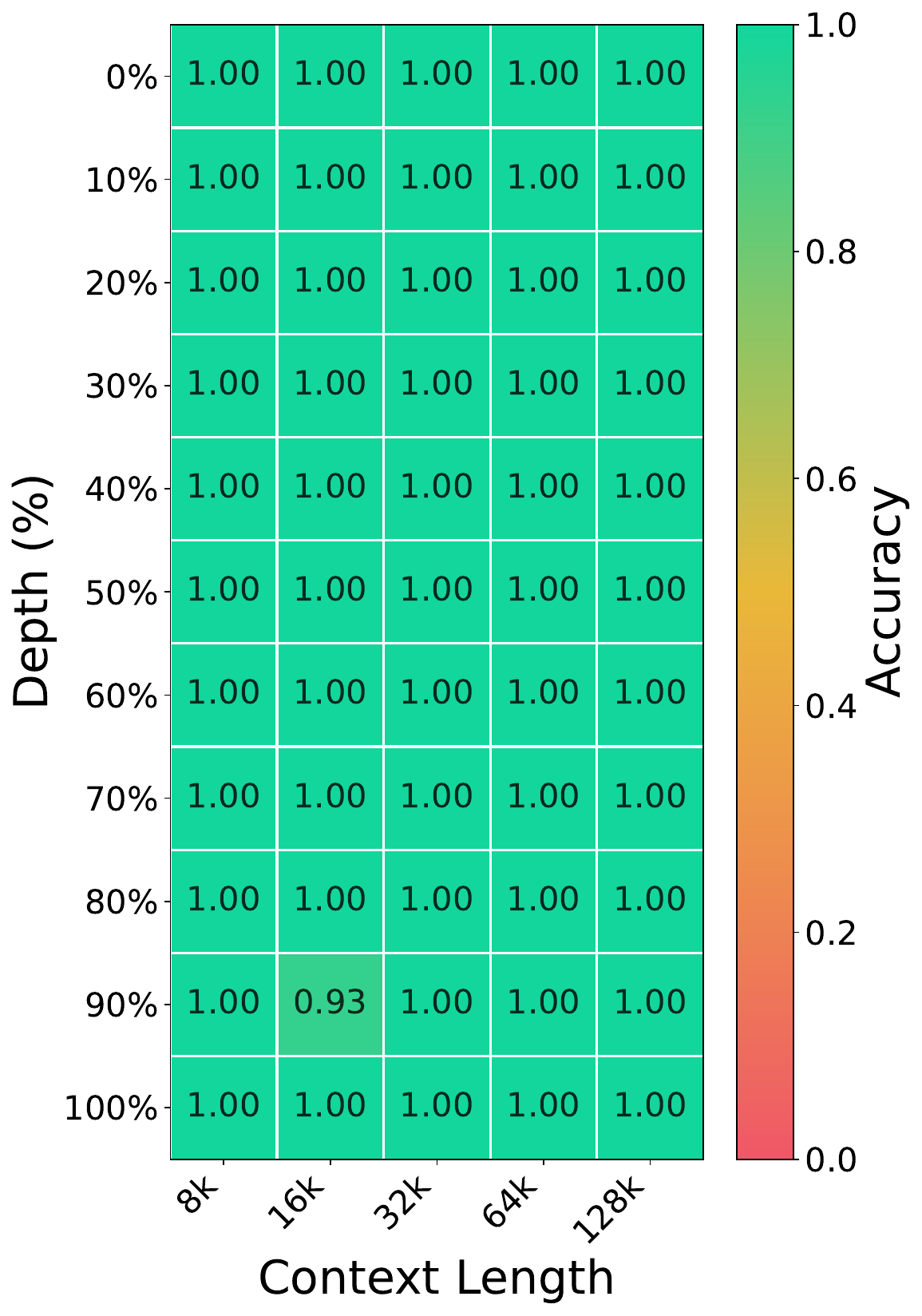}
        \caption{$h = 16$}
        \label{fig:nhead_stage2_h16}
    \end{subfigure}
    \caption{Ablation on the number of active heads $h$ used by the \emph{coarse} routed stage of the hierarchical variant \methodShort$^\dagger$ on DeepSeek-V3.2 ($H^I = 64$, $B = 1024$). The routed stage selects an enlarged candidate set of $k' = 4k = 8192$ tokens, which is then re-ranked by the full $H^I = 64$-head DSA indexer to obtain the final $k = 2048$ tokens. Each panel is a NIAH retrieval-accuracy heatmap at 128K context, with context length on the $x$-axis and needle depth on the $y$-axis. In comparison to the single-stage \methodShort{} results presented in Figure~\ref{fig:nhead}, the hierarchical pipeline demonstrates markedly greater robustness to a reduced number of routing heads ($h$). A key finding is that configurations with $h = 1$ and $h = 2$ in this two-stage framework achieve performance parity with the single-stage \methodShort{} utilizing twice as many heads (i.e., $h=2$ and $h=4$, respectively), because the second stage can recover any tokens that the routed stage missed within the enlarged candidate set.}
    \label{fig:nhead_stage2}
\end{figure}

Figure~\ref{fig:nhead_stage2} repeats the same sweep for the hierarchical variant \methodShort$^\dagger$, where the routed pass with $h$ heads now selects an enlarged candidate set of size $k' = 4k = 8192$, and the full $H^I = 64$-head DSA indexer then re-ranks this candidate set to extract the final $k = 2048$ tokens. Because the routed stage only has to keep the relevant tokens \emph{inside} the candidate set rather than pinpoint the exact top-$k$, the DSA refinement can compensate for very aggressive head reduction: A key observation is that the hierarchical variant \methodShort$^\dagger$ with $h = 1$ achieves comparable quality to the single-stage \methodShort{} with $h = 2$. Similarly, \methodShort$^\dagger$ with $h = 2$ matches the performance of the single-stage variant with $h = 4$. Even $h = 4$ delivers strong results at the full 128K context length. The hierarchical pipeline is therefore highly tolerant to under-routing in stage one, which is exactly the regime in which \methodShort$^\dagger$ achieves the strongest indexer fidelity (cf.\ Figure~\ref{fig:iou}) while still using a small $h$ in the heavy token-level scan.
\newpage
\section{Ablation: routing score}
\label{app:exp_metric}

\begin{figure}[htb]
    \centering
    \begin{subfigure}[t]{0.31\textwidth}
        \centering
        \includegraphics[width=\linewidth]{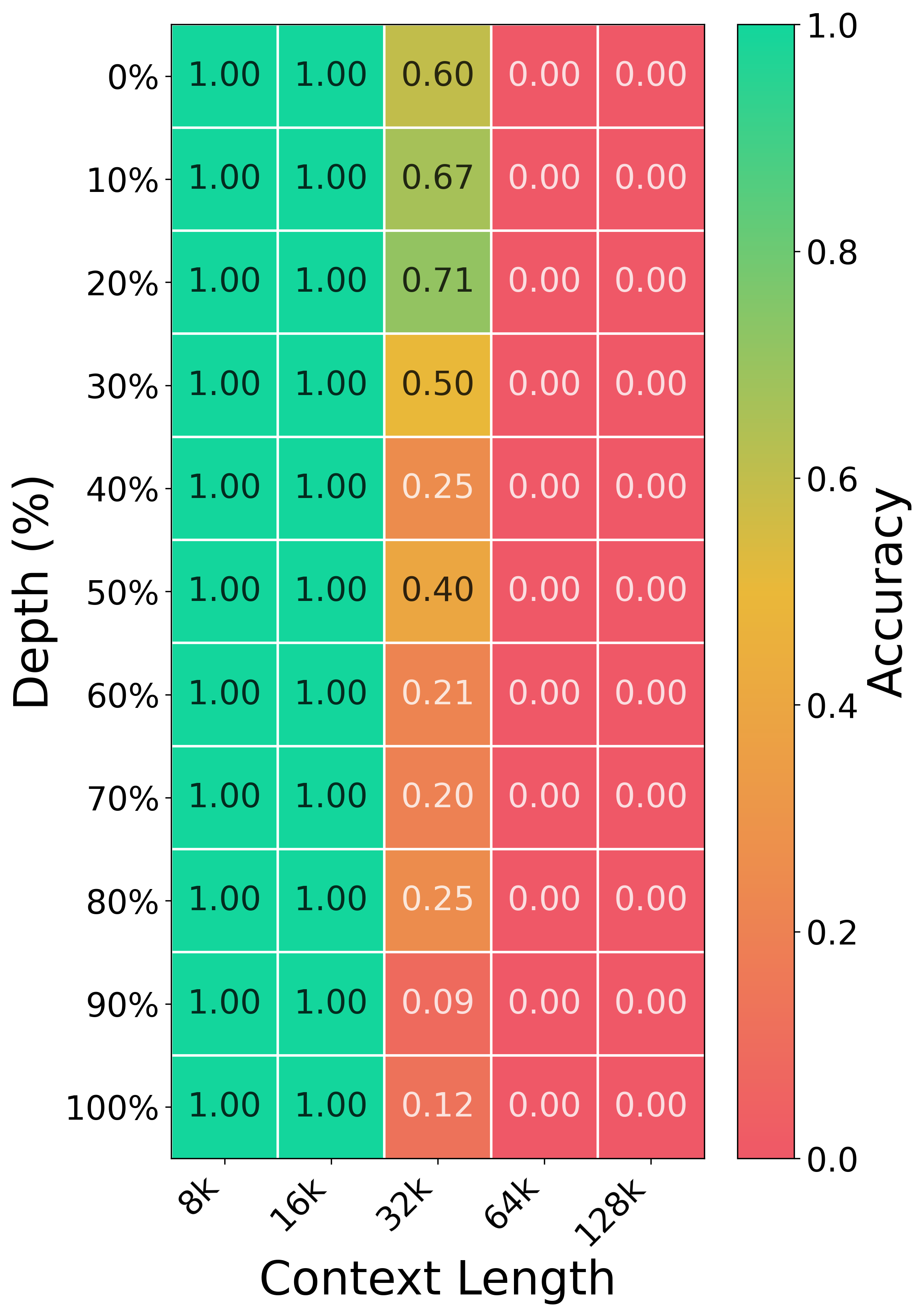}
        \caption{$w_{t,j}^I$}
        \label{fig:metric_weight}
    \end{subfigure}
    \hfill
    \begin{subfigure}[t]{0.31\textwidth}
        \centering
        \includegraphics[width=\linewidth]{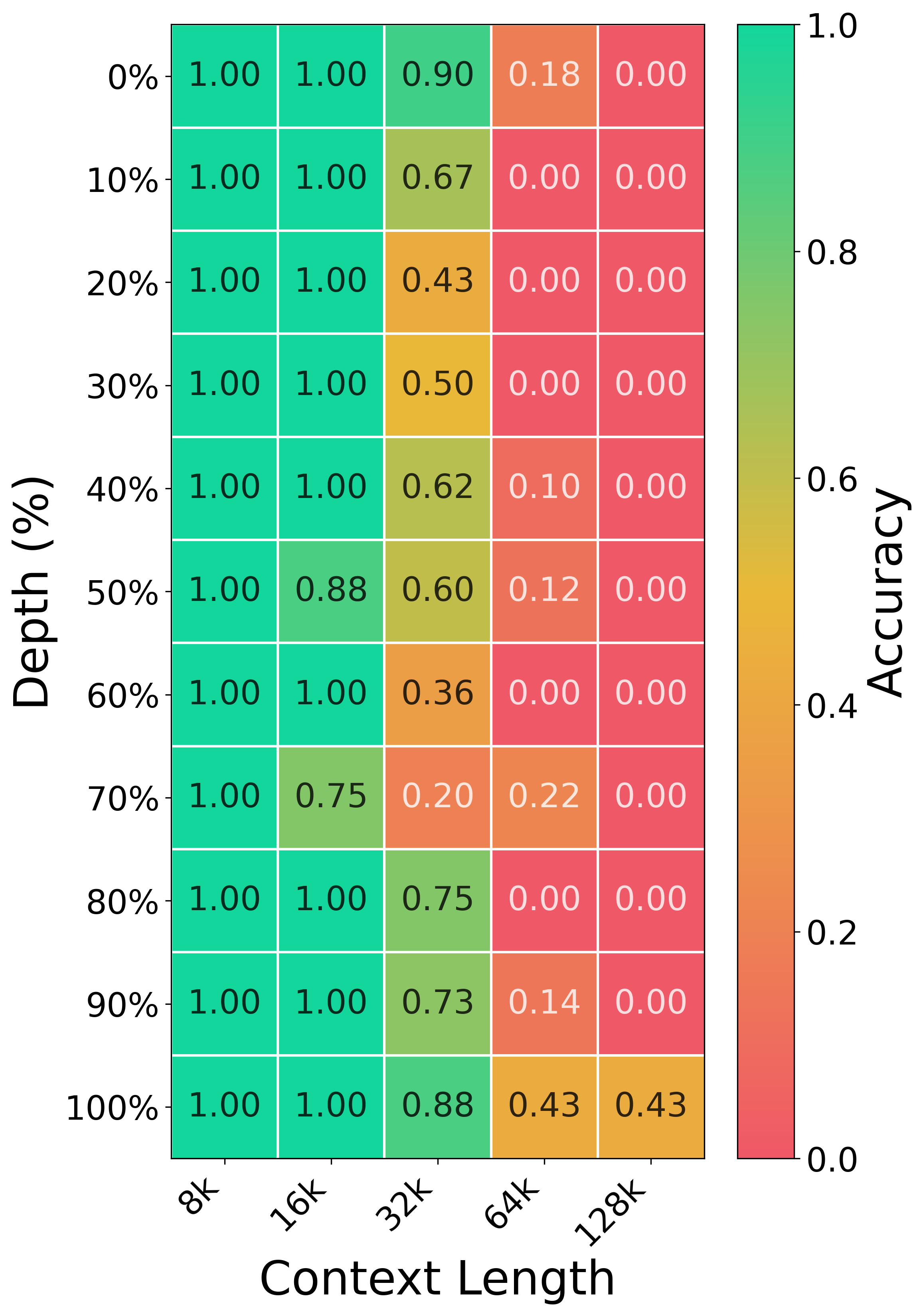}
        \caption{$\|\mathbf{q}_{t,j}^I\|_2$}
        \label{fig:metric_qnorm}
    \end{subfigure}
    \hfill
    \begin{subfigure}[t]{0.31\textwidth}
        \centering
        \includegraphics[width=\linewidth]{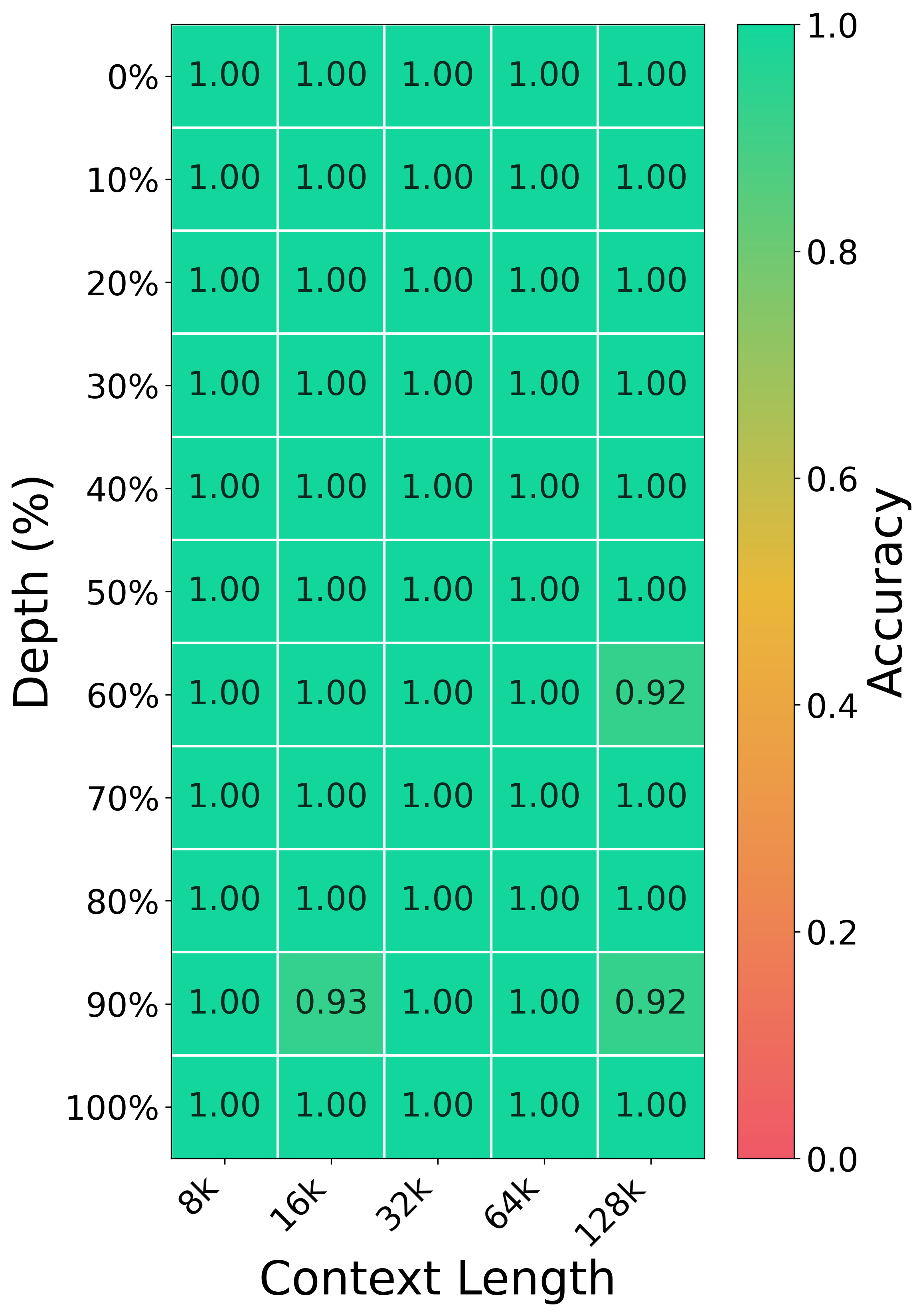}
        \caption{$\tfrac{1}{M}\sum_b \vert w_{t,j}^I\,\mathrm{ReLU}(\mathbf{q}_{t,j}^I\!\cdot\!\tilde{\mathbf{k}}_b^I)\vert $}
        \label{fig:metric_blockattn}
    \end{subfigure}
    \caption{Ablation on the head-importance score $E_{t,j}$ used by the \methodShort{} router on DeepSeek-V3.2 (Needle-in-a-Haystack accuracy at 128K). (a) the indexer gating weight alone; (b) the $\ell_2$ norm of the query head; (c) the proposed block-attention score, averaged over the $M$ pooled blocks of the prefix --- the only variant that actually consults past content. The $x$-axis denotes context length and the $y$-axis the needle depth (0\%--100\%); greener is better.}
    \label{fig:metric}
\end{figure}

The router score $E_{t,j}$ in Eq.~\ref{eq:misa_route_score} is the only learning-free signal that decides which $h$ heads are active for query $t$, so its choice is a critical design knob. We compare three candidates on DeepSeek-V3.2 with the default \methodShort{} setting ($h = 8$, $B = 1024$, $k = 2048$) and report NIAH accuracy at 128K (Figure~\ref{fig:metric}). The first two are content-light proxies that can be computed without ever consulting the prefix: (a) using the gating weight $w_{t,j}^I$ alone selects the heads that the DSA aggregator already up-weights; (b) the $\ell_2$ norm $\|\mathbf{q}_{t,j}^I\|_2$ picks the heads whose query directions are most ``confident''; None of these signals depends on what is actually \emph{in} the prefix that needs to be retrieved, and all two collapse on the harder regions of the NIAH grid. Variant (c), the proposed block-pooled attention $E_{t,j} = \tfrac{1}{M}\sum_b \vert w_{t,j}^I\,\mathrm{ReLU}(\mathbf{q}_{t,j}^I \cdot \tilde{\mathbf{k}}_b^I)\vert $, is the only score that aggregates query-to-prefix evidence into the routing decision, and is the only variant that fully recovers DSA's accuracy. This isolates the importance of \emph{where} the routing signal comes from rather than just \emph{which} heads are routed.

\newpage
\section{Ablation: router block size}
\label{app:exp_blocksize}
\begin{figure}[htb]
    \centering
    \includegraphics[width=0.7\linewidth]{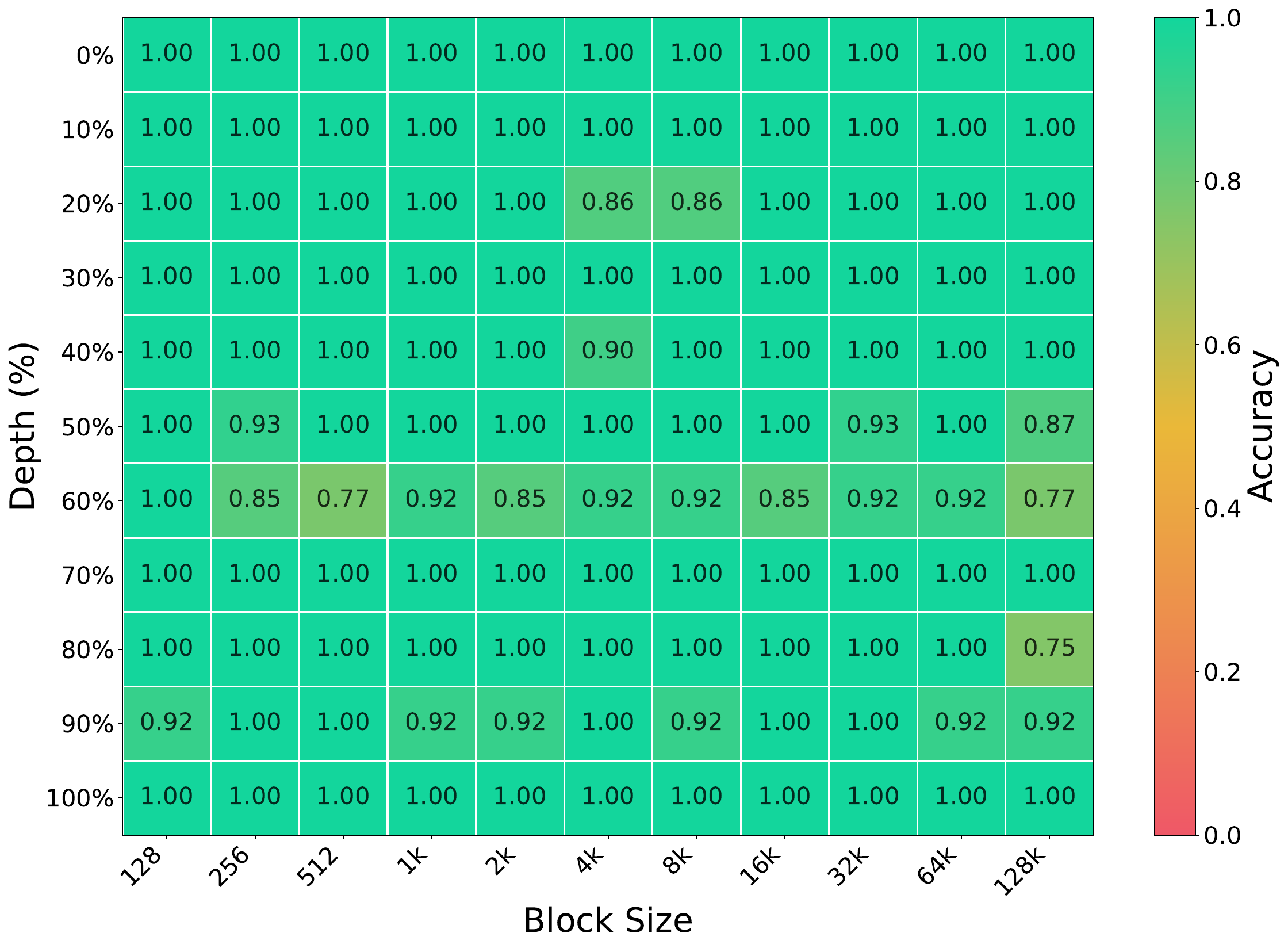}

    \caption{Ablation on the router block size $B$ on DeepSeek-V3.2 ($h = 8$, $k = 2048$). Each panel is a NIAH retrieval-accuracy heatmap at 128K context (context length on the $x$-axis, needle depth on the $y$-axis; greener is better). Accuracy is largely insensitive to $B$ across the full sweep, and only degrades mildly at the largest block sizes where the router is forced into a near-global pool.}
    \label{fig:blocksize}
    \end{figure}

Finally, we sweep the router block size $B$ on DeepSeek-V3.2 with $h = 8$ and $k = 2048$ fixed, ranging $B$ from $128$ to the full prefix length $L = 131{,}072$ (a single global pooled key)---eleven values, spanning three orders of magnitude. Recall that the router cost is $\mathcal{O}(H^I M)$ with $M = \lceil L / B \rceil$ pooled keys, so $B$ trades router cost against the spatial resolution of the head-importance estimate. Figure~\ref{fig:blocksize} shows the resulting NIAH heatmaps. Across this entire sweep, retrieval accuracy is largely insensitive to $B$: the heatmaps are visually indistinguishable from the dense DSA reference for the small to moderate block sizes, and only the very largest $B$ values (where the router is forced to summarise the prefix into a handful of pooled keys) begin to lose enough locality to introduce a mild degradation at the deepest needle depths. The default $B = 1024$ sits comfortably inside the stable region while keeping $M = 128$ pooled keys at $128$K context, which makes the router cost (a single $H^I \times M$ matmul per query) negligible compared to the subsequent token-level scoring.

\newpage
\section*{NeurIPS Paper Checklist}

The checklist is designed to encourage best practices for responsible machine learning research, addressing issues of reproducibility, transparency, research ethics, and societal impact. Do not remove the checklist: {\bf The papers not including the checklist will be desk rejected.} The checklist should follow the references and follow the (optional) supplemental material.  The checklist does NOT count towards the page
limit. 

Please read the checklist guidelines carefully for information on how to answer these questions. For each question in the checklist:
\begin{itemize}
    \item You should answer \answerYes{}, \answerNo{}, or \answerNA{}.
    \item \answerNA{} means either that the question is Not Applicable for that particular paper or the relevant information is Not Available.
    \item Please provide a short (1--2 sentence) justification right after your answer (even for \answerNA). 
\end{itemize}

{\bf The checklist answers are an integral part of your paper submission.} They are visible to the reviewers, area chairs, senior area chairs, and ethics reviewers. You will also be asked to include it (after eventual revisions) with the final version of your paper, and its final version will be published with the paper.

The reviewers of your paper will be asked to use the checklist as one of the factors in their evaluation. While \answerYes{} is generally preferable to \answerNo{}, it is perfectly acceptable to answer \answerNo{} provided a proper justification is given (e.g., error bars are not reported because it would be too computationally expensive'' or ``we were unable to find the license for the dataset we used''). In general, answering \answerNo{} or \answerNA{} is not grounds for rejection. While the questions are phrased in a binary way, we acknowledge that the true answer is often more nuanced, so please just use your best judgment and write a justification to elaborate. All supporting evidence can appear either in the main paper or the supplemental material, provided in appendix. If you answer \answerYes{} to a question, in the justification please point to the section(s) where related material for the question can be found.

IMPORTANT, please:
\begin{itemize}
    \item {\bf Delete this instruction block, but keep the section heading ``NeurIPS Paper Checklist"},
    \item  {\bf Keep the checklist subsection headings, questions/answers and guidelines below.}
    \item {\bf Do not modify the questions and only use the provided macros for your answers}.
\end{itemize}


\begin{enumerate}

\item {\bf Claims}
    \item[] Question: Do the main claims made in the abstract and introduction accurately reflect the paper's contributions and scope?
    \item[] Answer: \answerYes{} 
    \item[] Justification: The conclusions in the abstract and introduction align with the theoretical(see Section~\ref{method}) and experimental results(see Section~\ref{exp}).

\item {\bf Limitations}
    \item[] Question: Does the paper discuss the limitations of the work performed by the authors?
    \item[] Answer: \answerYes{} 
    \item[] Justification: See Section~\ref{limitation}.

\item {\bf Theory assumptions and proofs}
    \item[] Question: For each theoretical result, does the paper provide the full set of assumptions and a complete (and correct) proof?
    \item[] Answer: \answerYes{} 
    \item[] Justification: All the assumptions and proof are provided in Section~\ref{method}.

    \item {\bf Experimental result reproducibility}
    \item[] Question: Does the paper fully disclose all the information needed to reproduce the main experimental results of the paper to the extent that it affects the main claims and/or conclusions of the paper (regardless of whether the code and data are provided or not)?
    \item[] Answer: \answerYes{} 
    \item[] Justification: We provide detailed experimental results across various aspects, including LongBench(see Section~\ref{sec:exp_longbench}),NIAH(see Section~\ref{sec:exp_niah}), kernel speed experiments(see Section~\ref{sec:exp_speed}), ablation studies on the number of heads(see Section~\ref{sec:exp_nhead} and Appendix~\ref{app:exp_nhead2}), IoU(see Appendix~\ref{app:exp_iou}), score(see Appendix~\ref{app:exp_metric}) and block size(see Appendix~\ref{app:exp_blocksize}).

\item {\bf Open access to data and code}
    \item[] Question: Does the paper provide open access to the data and code, with sufficient instructions to faithfully reproduce the main experimental results, as described in supplemental material?
    \item[] Answer: \answerYes{} 
    \item[] Justification: The code will be released promptly after the paper submission. All datasets are publicly available.

\item {\bf Experimental setting/details}
    \item[] Question: Does the paper specify all the training and test details (e.g., data splits, hyperparameters, how they were chosen, type of optimizer) necessary to understand the results?
    \item[] Answer: \answerYes{} 
    \item[] Justification: Section~\ref{exp} already includes the models, datasets, hyperparameters, and hardware information used in the experiments to facilitate reproducibility.

\item {\bf Experiment statistical significance}
    \item[] Question: Does the paper report error bars suitably and correctly defined or other appropriate information about the statistical significance of the experiments?
    \item[] Answer: \answerNo{} 
    \item[] Justification: Due to prohibitive computational costs, detailed statistical information and corresponding error bars are omitted from the experimental results presented in this work.

\item {\bf Experiments compute resources}
    \item[] Question: For each experiment, does the paper provide sufficient information on the computer resources (type of compute workers, memory, time of execution) needed to reproduce the experiments?
    \item[] Answer: \answerYes{} 
    \item[] Justification: See the begining of Section~\ref{exp}.
    
\item {\bf Code of ethics}
    \item[] Question: Does the research conducted in the paper conform, in every respect, with the NeurIPS Code of Ethics \url{https://neurips.cc/public/EthicsGuidelines}?
    \item[] Answer: \answerYes{} 
    \item[] Justification: We have carefully reviewed and ensured that all aspects of our research
conform fully to the NeurIPS Code of Ethics.

\item {\bf Broader impacts}
    \item[] Question: Does the paper discuss both potential positive societal impacts and negative societal impacts of the work performed?
    \item[] Answer: \answerYes{} 
    \item[] Justification: See Section~\ref{conclusion}.
    
\item {\bf Safeguards}
    \item[] Question: Does the paper describe safeguards that have been put in place for responsible release of data or models that have a high risk for misuse (e.g., pre-trained language models, image generators, or scraped datasets)?
    \item[] Answer: \answerNA{} 
    \item[] Justification: The paper poses no such risks.

\item {\bf Licenses for existing assets}
    \item[] Question: Are the creators or original owners of assets (e.g., code, data, models), used in the paper, properly credited and are the license and terms of use explicitly mentioned and properly respected?
    \item[] Answer: \answerYes{} 
    \item[] Justification: we have carefully reviewed and verified the licenses and terms of use for all
existing assets used in this paper, including code, datasets, and pretrained models. The original creators are properly credited with citations to the corresponding papers.

\item {\bf New assets}
    \item[] Question: Are new assets introduced in the paper well documented and is the documentation provided alongside the assets?
    \item[] Answer: \answerYes{} 
    \item[] Justification: The code and environment configuration documentation will be released concurrently with the publication of the paper.

\item {\bf Crowdsourcing and research with human subjects}
    \item[] Question: For crowdsourcing experiments and research with human subjects, does the paper include the full text of instructions given to participants and screenshots, if applicable, as well as details about compensation (if any)? 
    \item[] Answer: \answerNA{} 
    \item[] Justification: The paper does not involve crowdsourcing nor research with human subjects.

\item {\bf Institutional review board (IRB) approvals or equivalent for research with human subjects}
    \item[] Question: Does the paper describe potential risks incurred by study participants, whether such risks were disclosed to the subjects, and whether Institutional Review Board (IRB) approvals (or an equivalent approval/review based on the requirements of your country or institution) were obtained?
    \item[] Answer: \answerNA{} 
    \item[] Justification: The paper does not involve crowdsourcing nor research with human subjects.

\item {\bf Declaration of LLM usage}
    \item[] Question: Does the paper describe the usage of LLMs if it is an important, original, or non-standard component of the core methods in this research? Note that if the LLM is used only for writing, editing, or formatting purposes and does \emph{not} impact the core methodology, scientific rigor, or originality of the research, declaration is not required.
    \item[] Answer: \answerNA{} 
    \item[] Justification: LLM is not an important, original, or non-standard component of the core methods in this research.

\end{enumerate}

\end{document}